\documentclass[a4paper,fleqn]{cas-dc} 

\usepackage[compress]{natbib}
\usepackage{amsmath}

\usepackage{times}
\usepackage{epsfig}
\usepackage{graphicx}
\usepackage{amsmath}
\usepackage{amssymb}
\usepackage{amsthm}
\usepackage{booktabs}
\usepackage{multirow}
\usepackage{pifont}
\usepackage{booktabs} 
\usepackage{multirow}
\usepackage{svg}
\usepackage{xcolor} 
\usepackage{amsmath}

\usepackage{pifont}%
\usepackage{newunicodechar}
\newunicodechar{≥}{\ensuremath{\ge}}

\usepackage{graphicx}
\usepackage{float}
\usepackage{hyperref}
\usepackage{subcaption}
\usepackage{diagbox}

\usepackage{algorithm}
\usepackage{algpseudocode}
\usepackage{tabularx}
\usepackage{comment}
\usepackage{xcolor}
\usepackage[export]{adjustbox}

\usepackage[acronym, nonumberlist, nopostdot]{glossaries}
\makeglossaries
\newacronym{AI}{AI}{Artificial Intelligence}
\newacronym{API}{API}{Application Programming Interface}
\newacronym{APoZ}{APoZ}{Average Percentage of Zeroes}
\newacronym{ARM}{ARM}{Advanced RISC Machines}
\newacronym{ASICs}{ASICs}{ Application Specific Circuits}
\newacronym{ASC}{ASC}{Application Specific Circuits}
\newacronym{AXI}{AXI}{Advanced eXtensible Interface}
\newacronym{BN}{BN}{Batch Normalization}
\newacronym{BNNs}{BNNs}{Binarized Neural Networks}
\newacronym{CAD}{CAD}{Computer Aided Design}
\newacronym{CAM}{CAM}{Camera}
\newacronym{CLE}{CLE}{Cross Layer Equalization}
\newacronym{CMSIS}{CMSIS}{Cortex Microcontroller Software Interface Standard}
\newacronym{CNN}{CNN}{Convolutional Neural Networks}
\newacronym{CPU}{CPU}{Central Processing Unit}
\newacronym{DL}{DL}{Deep Learning}
\newacronym{DPU}{DPU}{Deep Learning Processor Unit}
\newacronym{DSPs}{DSPs}{Digital Signal Processors}
\newacronym{FINN}{FINN}{Fast, Scalable Quantized Neural Network Inference on FPGAs}
\newacronym{FLOPs}{FLOPs}{Floating-Point Operations}
\newacronym{FPGAs}{FPGAs}{Field Programmable Gate Arrays}
\newacronym{FPS}{FPS}{Frames Per Second}
\newacronym{GPU}{GPU}{Graphics Processing Unit}
\newacronym{JI}{DI}{Dice Index}
\newacronym{HAL}{HAL}{Hardware Abstraction Layer}
\newacronym{HDMI}{HDMI}{High-Definition Multimedia Interface}
\newacronym{HW}{HW}{Hardware}
\newacronym{IoT}{IoT}{Internet of Things}
\newacronym{IoU}{IoU}{Intersection Over Union}
\newacronym{IP}{IP}{Intellectual Property}
\newacronym{IR}{IR}{Infrared}
\newacronym{LoRa}{LoRa}{Long Range}
\newacronym{MACs}{MACs}{Multiply-Accumulate Operations}
\newacronym{MAPE}{MAPE}{Mean Absolute Percentage Error}
\newacronym{ML}{ML}{Machine Learning}
\newacronym{MPSoC}{MPSoC}{Multiprocessor System-on-Chip}
\newacronym{NN}{NN}{Neural Network}
\newacronym{NPU}{NPU}{Neural Processor Unit}
\newacronym{NRE}{NRE}{Non-recurring Engineering}
\newacronym{OS}{OS}{Operating System}
\newacronym{PE}{PE}{Processing Element}
\newacronym{PL}{PL}{Programmable Logic}
\newacronym{PS}{PS}{Processing System}
\newacronym{ReLU}{ReLU}{Rectified Linear Unit}
\newacronym{RMSPE}{RMSPE}{Root Mean Square Percentage Error}
\newacronym{SGDW}{SGDW}{Stochastic Gradient Descent with Decoupled Weight}
\newacronym{SoC}{SoC}{System-on-Chip}
\newacronym{TfT}{TfT}{Thin-Film Transistor}
\newacronym{UNet}{UNet}{Unified Network}
\newacronym{USB}{USB}{Universal Serial Bus}
\newacronym{VART}{VART}{Vitis AI Run Time}
\newacronym{VPU}{VPU}{Vision Processing Unit}

\def\tsc#1{\csdef{#1}{\textsc{\lowercase{#1}}\xspace}}
\tsc{WGM}
\tsc{QE}
\tsc{EP}
\tsc{PMS}
\tsc{BEC}
\tsc{DE}

\begin{document}
\let\WriteBookmarks\relax
\def\floatpagepagefraction{1}
\def\textpagefraction{.001}

\shorttitle{A reconfigurable smart camera implementation for jet flames characterization based on BNNs}

\shortauthors{G.V. Vazquez-Garcia}

\title [mode = title]{A reconfigurable smart camera implementation for jet flames
characterization based on an optimized segmentation model}

\author[inst1]{Gerardo Valente Vazquez-Garcia}
\author[inst2]{Carmina Perez Guerrero}
\author[inst2]{Eduardo Garduño}
\author[inst2]{Miguel Gonzalez-Mendoza}
\author[inst3]{Adriana Palacios}
\author[inst2]{Gerardo Rodriguez-Hernandez} 
\author[inst4]{Vahid Foroughi}
\author[inst4]{Alba Àgueda}
\author[inst4]{Elsa Pastor}
\author[inst2]{Gilberto Ochoa-Ruiz}

\affiliation[inst1]{organization={Universidad Autónoma de Guadalajara, Maestria en Ciencias Computacionales},
    city={Guadalajara}, postcode={45017},
    state={Jalisco}, country={Mexico}}

\affiliation[inst2]{organization={Escuela de Ingeniería y Ciencias, Tecnológico de Monterrey},
    city={Monterrey}, postcode={64849},
    state={N.L.}, country={Mexico}}

\affiliation[inst3]{organization={Universidad de las Américas Puebla, Department of Chemical, Food and Environmental Engineering, Puebla, 72810, Mexico}}

\affiliation[inst4]{organization={Universitat Politècnica de Catalunya - BarcelonaTech. EEBE, Eduard Maristany 16, 08019 Barcelona. Catalonia, Spain}}

\maketitle

\begin{abstract}
In this work we present a novel framework for fire safety management in industrial settings through the implementation of a smart camera platform for jet flames characterization. The proposed approach seeks to alleviate the lack of real-time solutions for industrial early fire segmentation and characterization. As a case study, we demonstrate how a \gls{SoC} \gls{FPGAs}-based platform, running optimized \gls{AI} models can be leveraged to implement a full edge processing pipeline for jet flames analysis. Jet flames are one of the less studied sources of major accidents. In this paper we extend previous work on computer-vision jet fire segmentation by creating a novel experimental set-up and system implementation for addressing this issue, which can be replicated to other fire safety applications. The proposed platform is designed to carry out image processing tasks in real-time and on device, reducing video processing overheads, and thus the overall latency. This is achieved by optimizing a UNet segmentation model to make it amenable for an SoC FPGAs implementation; the optimized model can then be efficiently mapped onto the SoC reconfigurable logic for massively parallel execution. For our experiments, we have chosen the Ultra96 platform, as it also provides the means for implementing full-fledged intelligent systems using the SoC peripherals, as well as other \gls{OS} capabilities (i.e., multi-threading) for systems management. For optimizing the model we made use of the Vitis (Xilinx) framework, which enabled us to optimize the full precision model from 7.5 million parameters to 59,095  parameters (125x less), which translated into a reduction of the processing latency of 2.9x. Further optimization (multi-threading and batch normalization) led to an improvement of 7.5x in terms of latency, yielding a performance of 30 \gls{FPS} without sacrificing accuracy in terms of the evaluated metrics (Dice Score). We also show that the proposed architecture and model can effectively produce the desired characterization of horizontal jet flames, attaining performance similar to that of the full precision model.



\end{abstract}

\section{Introduction}

\begin{figure*}[h]
\centering
\includegraphics[width=0.7\linewidth]{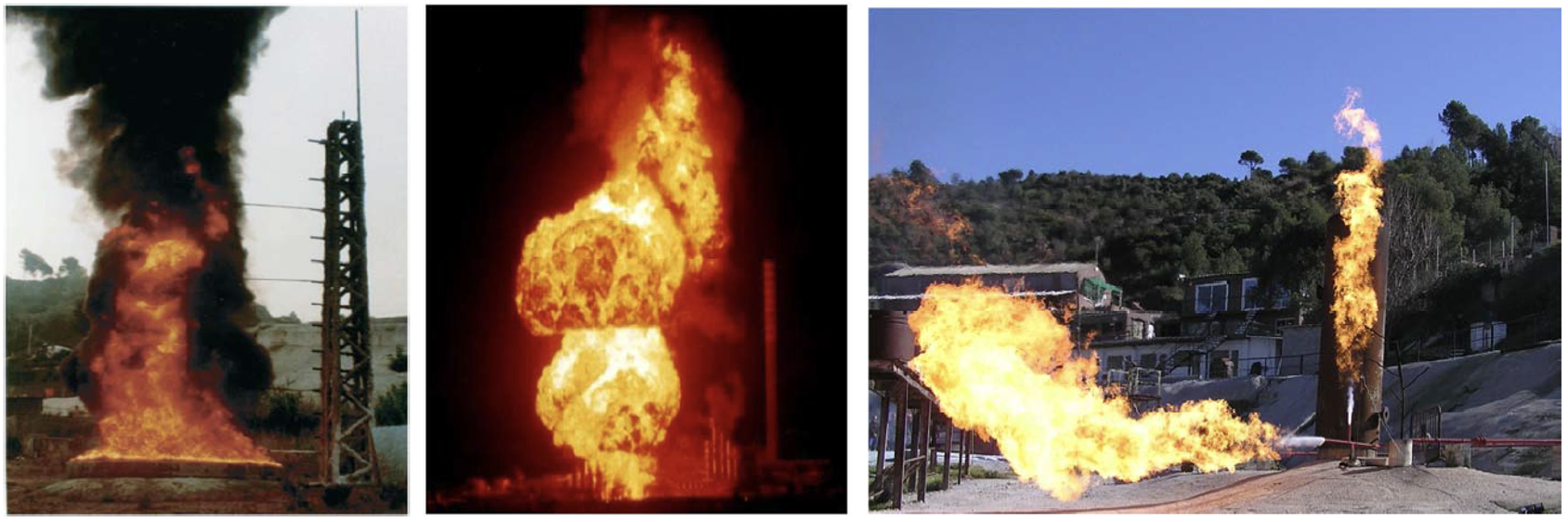}
    \caption{Different types of flames patterns in industrial settings: pool, ball and jet fires. Modified from Palacios (2011).}
\label{img:typesFlames}
\end{figure*}

\begin{figure*}[t]
\centering
\includegraphics[width=0.9\linewidth]{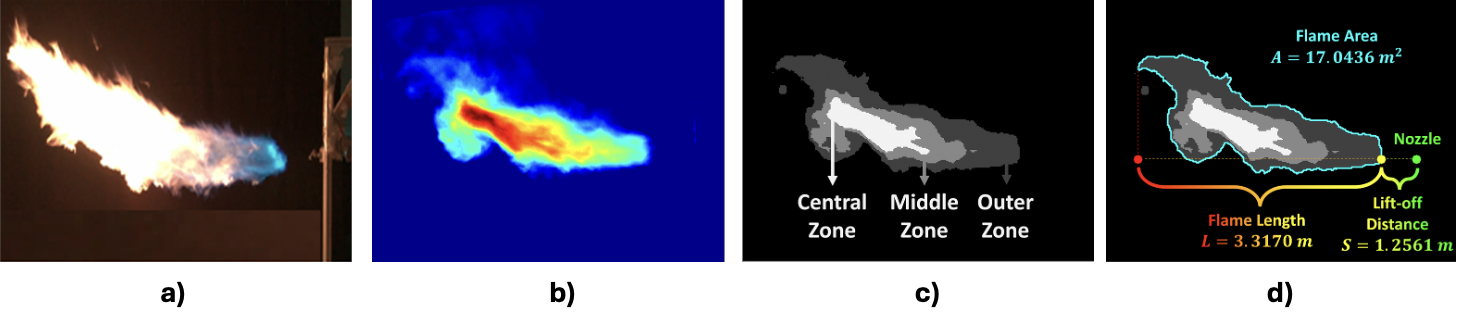}
\vspace{4pt}
    \caption{Visual example of jet fire characterization based on segmentation of three temperature zones.}
\label{img:firecharacteristics}
\end{figure*}

The production, handling and transport of hazardous materials in the chemical industry have increased during the last decades leading to a higher frequency of major accidents. The consequences that these have had on people and the environment have highlighted the need to improve industrial safety measures. This has been reflected in the implementation of legislation such as the Seveso Guideline on the prevention of major accidents in certain industrial activities \cite{OJEC2012}. Thus, the development of research studies on the main characteristics, mathematical modeling, and estimation of consequences of these major accidents occurring in the chemical industry around the world is of paramount importance.

Fires are one of the most common and serious threats to the safety of both equipment and people operating in process units. 
Among the various types of major fires that can occur (see Fig. \ref{img:typesFlames}), jet fires (Fig. \ref{img:firecharacteristics} (a) and (b)), are one of the lesser-known types, something that has changed recently, with a surging interest in the research community due to safety concerns for high-pressure storage and transport of fuels like hydrogen. 
Jet fires are a particular type of serious major fire accident that can
occur both in fixed installations (process plants, storage,
or pipes) and, less frequently, during transport
(trucks or tank cars). Although jet fires occur with some
frequency, many of them are not reported, since they are
often accompanied by other events of greater magnitude,
such as explosions. 

Jet fires can be evaluated based on specific geometrical features (Fig. \ref{img:firecharacteristics} (c) and (d)). This is relevant from a risk management perspective because the probability of domino effect increases due to flame impingement, for example. Also, being able to predict the shape and proportions of a jet fire is relevant since it is possible then to identify the optimal space for the location of equipment and structures.

A great majority of consequence analyses regarding jet fires,
that evaluate both geometric features and radiation loads,
employ physics-based models, such as Computational
Fluid Dynamics (CFD) \cite{colella2020}, or empirical models, which are
based on correlations developed from a set of experiments. However, few works have tackled this problem from a data-centric perspective. Furthermore, most empirical models use a simplified
flame geometry and a fixed emissive power to determine the
thermal radiation load on a target.

Research done so far to characterize the main geometrical features of jet flames include the work from \cite{fuel2021}, which presents a systematic study of the factors that affect the lift-off distance, which is the length between the burner exit and the base of the lifted flame. It was found that the lift-off distance varies linearly according to the exit velocity, and it is independent of the burner diameter. \cite{palacios11} also analyzed the shape, length, and width of relatively large vertical jet fires at sonic and subsonic exit velocities. The flame boundary was defined at a temperature of 800 K, and the results indicated that a cylindrical shape could accurately describe the shape, length, and diameter of the flame. Additionally, \cite{Bradley16} introduced an extensive review and re-thinking of jet flame heights and structure, including a proposal of dimensionless correlations for the atmospheric jet flames' heights and lift-off distances. It was found that the same flow rate parameter could be used to correlate both heights and flame lift-off distances, and based on that, an equation was proposed to predict jet flame height including several fuels and orientations.

More recently, there has been a surging interest in using \gls{ML} techniques to model jet fires. For instance, \cite{ Lattimer20} presented an overview of \gls{ML}
techniques for low-cost and high-fidelity fire predictions where it was found that \gls{ML} has the capacity to provide full-field predictions faster than CFD simulations. In this sense, a recent research work \citep{carmina_journal} has  explored the use of \gls{DL}-based computer vision methods for jet fires segmentation and characterization. Those methods were conceived for offline applications and made use of models deployed in cloud applications (i.e., they were not optimized for edge applications such as smart sensors). However, since monitoring jet fires is a critical task to provide a timely and adequate response, \gls{ML} models must be capable of detecting and characterizing such situations preferably as smart sensors providing real-time alerts. This entails that these devices should be in close proximity to the areas where jet fires can potentially occur. 


As a solution for video processing in industrial \gls{IoT} applications, several researchers and practitioners have proposed the use of smart cameras \citep{smart_camera}, which main purpose is to implement all the functionalities of a conventional camera system (i.e., image acquisition, storage and video outputting) with an important difference: such smart devices are capable of performing image processing at the edge. The main rationale for this approach is to reduce the amount of data to be transmitted by performing feature extraction or carrying out specific tasks on-board, reducing in this manner the required throughput and being less affected by latency.

In particular, recent works have shown the integration of \gls{DL} models, optimized in such a way that they can be used to perform inference in embedded devices \citep{BNN_camera}. In this work, we propose a smart camera implementation for jet fire characterization using a UNet architecture, which has been optimized for running on an \gls{SoC} \gls{FPGAs} platform. In particular, we are interested in jet fire characterization in-situ and on-device, which entails that the jet flame is automatically segmented into three zones (Fig. \ref{img:firecharacteristics} (c)), whilst  estimating the geometrical characteristics (Fig.  \ref{img:firecharacteristics} (d)).

\subsection{Objectives}

The main objective of this work is to demonstrate a novel approach for implementing real-time image processing and analysis for smart cameras. The specific objectives are as follows: 

\begin{itemize}
    \item To propose a framework for implementing \gls{BNNs}-based computer vision algorithms in an \gls{SoC} \gls{FPGAs} platform targeted to smart cameras implementations.
    \item To introduce a customized segmentation model based on UNet which is tailored for a specific case study, i.e., real-time jet flames categorization into three main zones and further geometrical characterization. 
    \item To create a proof of concept of  an adaptable and extensible architecture for fire monitoring that can be used for other applications beyond the case study presented herein.
    
\end{itemize}

The rest of this paper is organized as follows: 
Section \ref{State of the Art} surveys the state of the art and the motivation for implementing optimized neural networks on constrained devices. Section \ref{Methodology} describes the methodology for implementing our solution, and Sections \ref{Results} and  \ref{Discussion} include a description of our results and a brief discussion providing recommendations for future investigation. Section \ref{Conclusion} summarizes the findings.

\begin{figure*}[h]
\centering
\includegraphics[width=0.65\linewidth]{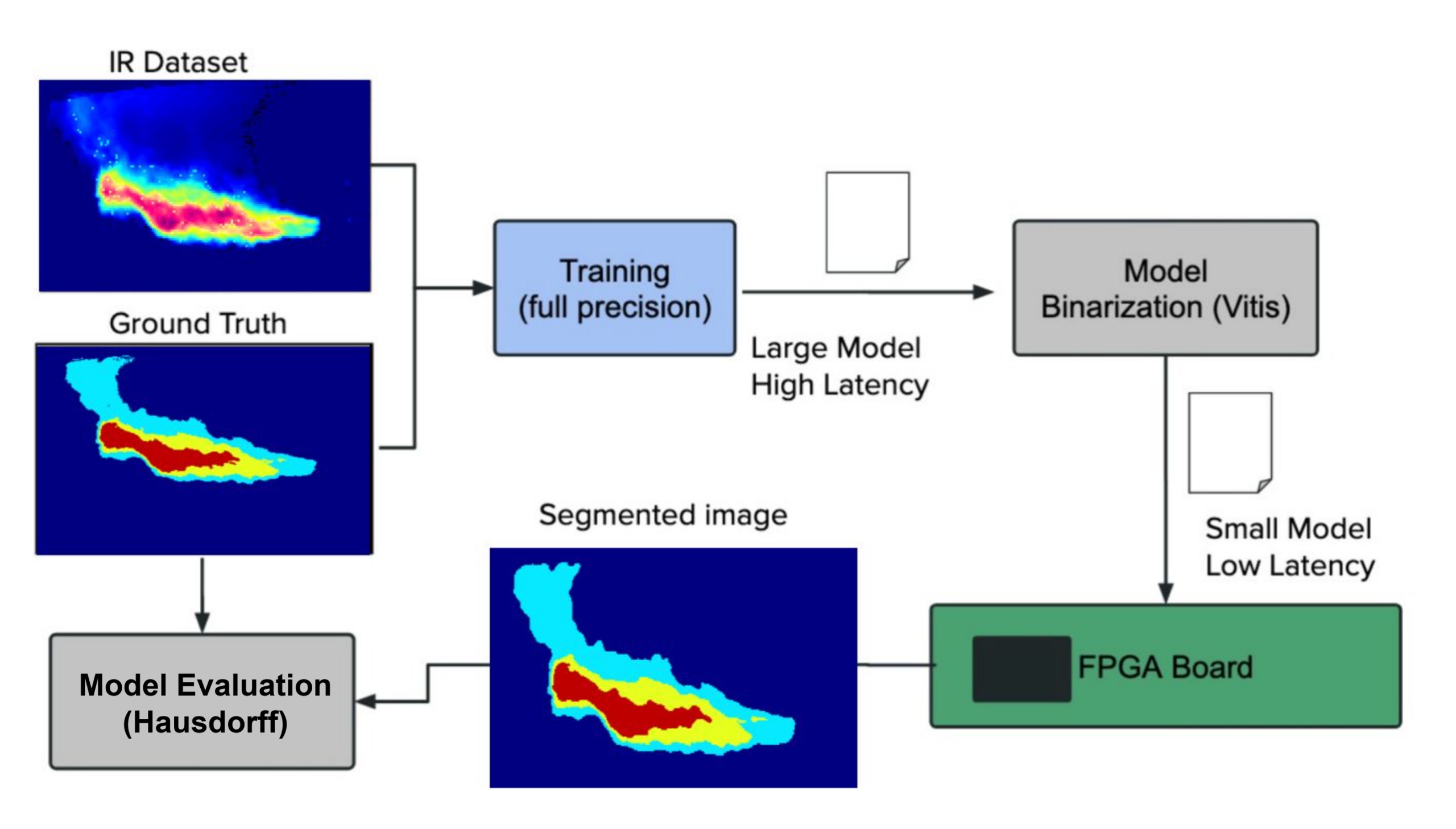}
    \caption{Overall flow followed in this work: a full precision model is trained using an \gls{IR} dataset of jet flames. The model is then optimized using Vitis and the resulting moddel is ported on an \gls{FPGAs}-based \gls{SoC} architecture for evaluation (adapted from \cite{gardunno}).}
\label{img:SoCFPGAgral}
\end{figure*}

\section{State of the art}
\label{State of the Art}

\subsection{Cloud vs edge}

Cloud computing is defined as a \textit{"model for enabling ubiquitous, convenient, on-demand network access to a shared pool of configurable computing resources (e.g., networks, servers, storage, applications, and services) that can be rapidly provisioned and released with minimal management effort or service provider interaction"} \citep{motiv_mell_2011}. At first glance cloud computing seems to be a good paradigm to be used for critical tasks, such as jet fires monitoring. In practice, however, there are several characteristics that make cloud-based solutions sub-optimal. For example \cite{motiv_sunyaev_2020} mentions that cloud computing applications can face communication issues either due to network issues or service denial, whereby the reliability of the whole system could be compromised.

\begin{figure*}[t]
\centering
\includegraphics[width=0.55\linewidth]{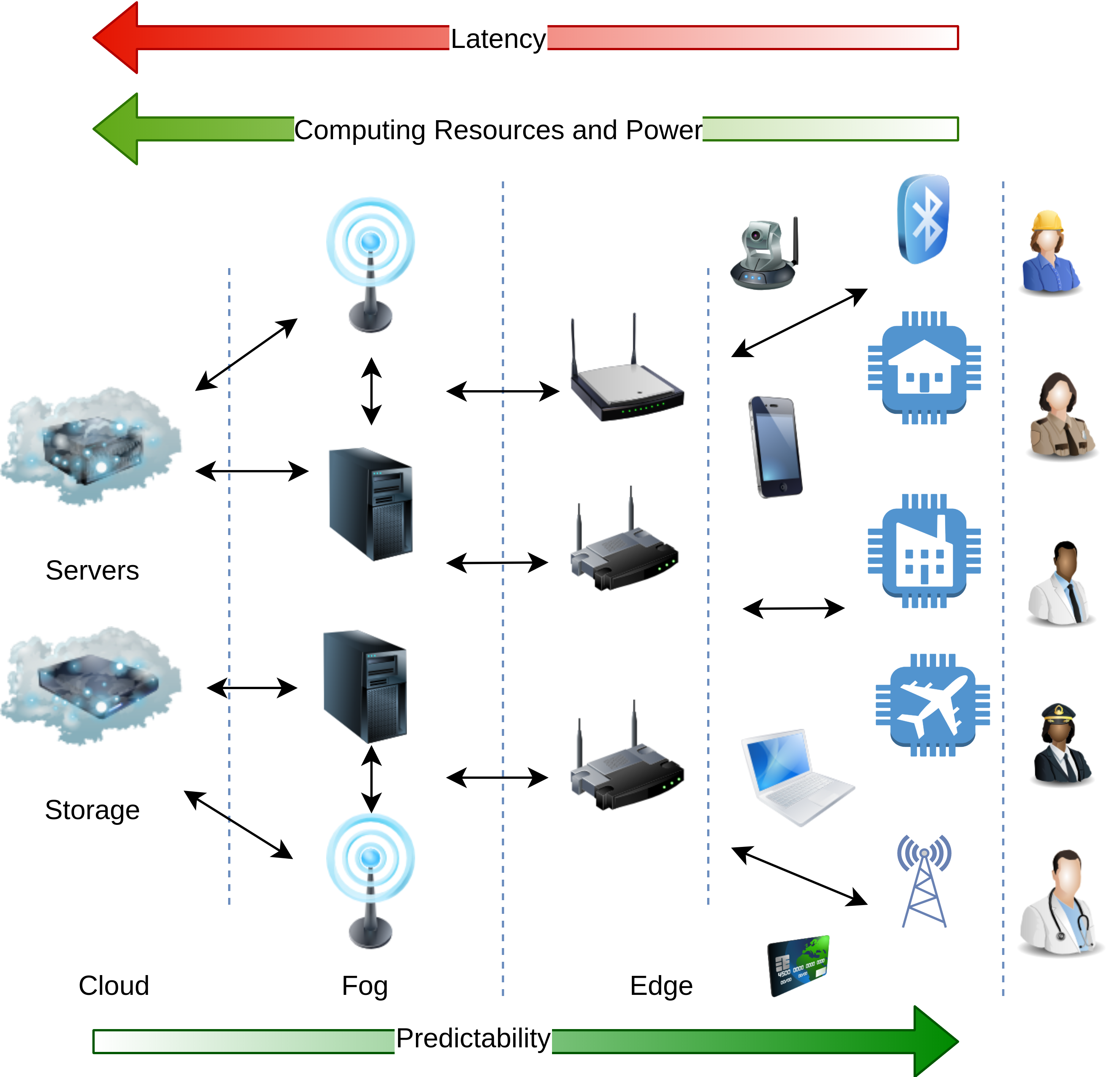}
    \caption{Edge computing vs cloud computing.}
\label{img:EdgeComp}
\end{figure*}

On the other hand, edge computing is a paradigm in which the processing of the information is done close to its source, on a so-called edge device. With this approach, there is no need to transfer the collected information through a wide network, such as the internet, and hence the latency between the information being captured and being processed decreases. At the same time, the predictability and reliability of the whole system increases \citep{better_predict}. However, one downside of the edge approach is that an edge device is limited in computing resources and power availability. Therefore, in order to deal with limited hardware and power resources, it is important to develop power efficient hardware architectures, suitable for different use cases. On this front, reconfigurable computing devices are good candidates, since they provide good performance at low power consumption levels while allowing the deployment of different programmable hardware implementations \citep{sota_vestias_2019}. Fig.  \ref{img:EdgeComp} shows some of the existing trade-offs between cloud and edge computing.

The above-mentioned issues are particularly pressing when \gls{AI} systems (in particular, computer vision systems) are deployed in industrial environments because the amount of data (images or video) is much larger than for other applications. Thus, the combination of high latency and high throughput makes it impossible to develop real-time monitoring systems based on image analysis algorithms, and new paradigms are required.

\gls{DL} approaches have shown groundbreaking  progress over the last two decades in a large number of applications. Once trained, 
\gls{AI} models based either on 
\gls{CNN} or transformers can then be deployed for inference for tasks such as image recognition, natural language processing, and speech recognition. These models can leverage the power of \gls{GPU}s in order to achieve higher on-device performance.

On-device \gls{AI} refers to the deployment of AI models that run directly on local hardware, also known as edge devices, instead of relying on cloud-based systems. This enables the device to process data locally, making decisions without having to send information to a central server for analysis. It contrasts with cloud-based \gls{AI}, which requires constant data transmission to and from external servers, introducing latency and privacy concerns. This computing model is increasingly being adopted across industries because it offers real-time insights and decision-making, especially in scenarios where immediate feedback is crucial. 

Edge \gls{AI} is optimized for running on the device’s available computing resources, such as \gls{CPU}s, \gls{GPU}s, or specialized \gls{AI} accelerators such as \gls{FPGAs} or application-specific integrated circuits. Thus, understanding \gls{CPU} utilization is crucial for efficient resource management, particularly during model conversion and integration processes, to achieve faster and more accurate inference. This ensures that models are designed to run efficiently without consuming excessive power or memory. The rise of energy-efficient processors and model optimization techniques such as pruning and quantization, further enhance the performance of \gls{AI} models on limited-resource devices like smartphones, wearables, and \gls{IoT} devices. As it will be discussed next, this capability ensures that devices with constrained processing power (i.e., smart cameras) can run \gls{AI} models more effectively. 

\begin{table*}
\caption{Types of smart cameras and their main application types.}
\label{tab: sota_smartcams_t}
\setlength{\tabcolsep}{3pt}
\begin{tabular}{|p{90pt}|p{280pt}|}
\hline
\textbf{Smart Camera type} & \textbf{Main Applications}\\
\hline
Single-chip & Information sensors.\\
Embedded & Optical mice, fingerprint readers, smartphones \\
Stand-alone & Industrial machine vision, human computer interfaces\\
Compact system & Security, traffic surveillance and machine vision\\
Distributed & Intelligent video surveillance, industrial machine vision, information gathering systems\\
\hline
\multicolumn{2}{p{230pt}}{}\\
\end{tabular}
\end{table*}

\begin{figure*}[h]
\centering
\includegraphics[width=0.7\linewidth]{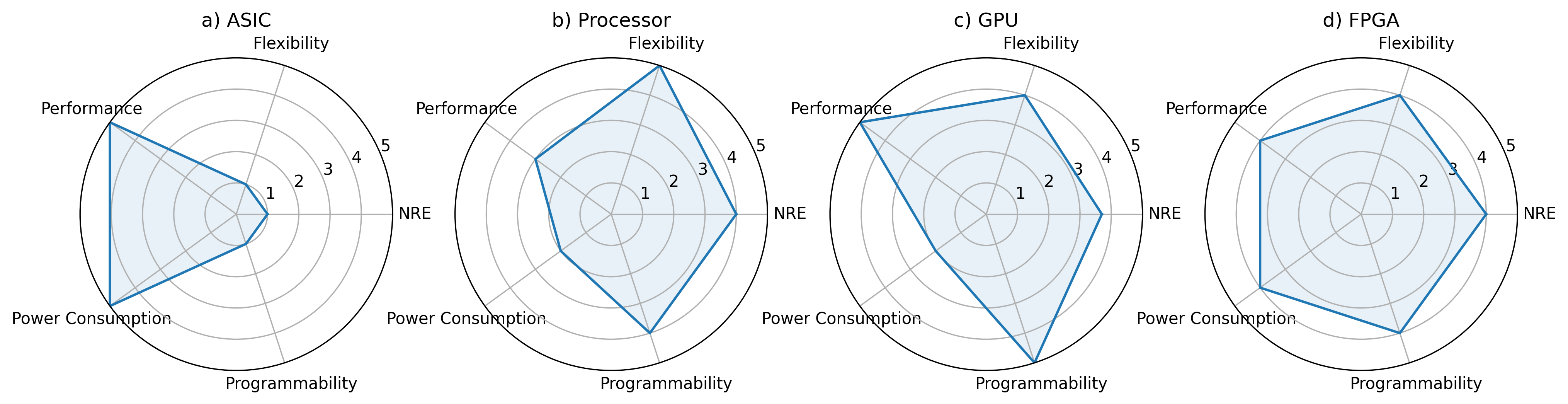}
    \caption{Comparison in terms of \gls{NRE}, flexibility, processing performance, power consumption and programmability of different technological choices for implementing smart cameras : a) \gls{ASICs}; b) Processor (i.e.,\gls{SoC}) and Digital Signal Processors ( \gls{DSPs}); c) \gls{GPU} and d)  \gls{FPGAs} (figure adapted from \cite{Real2010} )}

\label{img:ComputingLvls}
\end{figure*}

\subsection{Smart cameras}

Smart cameras are systems capable of processing, analyzing, and extracting information from the images they collect \citep{sota_shi_2009}. The extraction is done through different video processing algorithms. Smart cameras have been used in different applications such as human gesture recognition \citep{sota_wolf_2002}, surveillance \citep{sota_bramberger_2006}, smart traffic light optimization systems \citep{sota_tchuitcheu_2020}, and even a fire detection system based on traditional image processing techniques has been proposed by \citet{sota_gomes_2014}.  

Due to their capacity for complex pattern identification, smart cameras are good candidates for several types of applications. However, in order to achieve optimal performance, it is crucial to choose the right hardware for every application. As of now, devices such as microcontrollers, \gls{GPU}s, \gls{DSPs}, microprocessors, media processors, \gls{ASICs} and \gls{FPGAs} can be used for implementing smart cameras. Sometimes these devices have been combined on different architectures for such purpose \citep{sota_real_2009}. 

\subsection{Technological implementation choices}

When designing embedded smart systems such as edge \gls{AI} cameras, the selection of a particular hardware implementation can directly impact aspects such as cost, performance, power consumption and the configurability of the final system \citep{sota_vestias_2019}. Table \ref{tab: sota_smartcams_t} shows a classification of smart cameras based on the hardware device used, as proposed by \cite{sota_shi_2009}. 

As it can be observed in Fig. \ref{img:ComputingLvls} based on the analysis from \cite{sota_shi_2009}, different technological choices have impact on the adaptability and on system performance. It is important to note that none of these devices can possibly suit every application and there are different trade-offs and requirements to be met. As a matter of fact, \gls{ASICs} are devices with the highest performance and lowest energy consumption, but they are traditionally the most expensive solution (i.e., due to fabrication costs) and cannot be upgraded. \gls{CPU}s offer a high degree of reconfigurability but they are the slowest devices, as they do not offer parallelization capabilities. \gls{GPU}s, on the other hand, provide a high throughput, but in such devices the power consumption is prohibitive for many edge applications \citep{sota_vestias_2019}. Finally, \gls{FPGAs} devices have been traditionally used to create more power-efficient hardware accelerators capable of attaining high throughput \citep{sota_venieris_2018}, while maintaining a high flexibility and reconfigurability so that the final system can be upgraded to support new workloads or speed up the current ones via updates on the hardware design, all while maitaining a low-power consumption.

The downside of \gls{FPGAs} is that in order to reach these goals, developers require hardware design expertise \cite{ochoa}, therefore the design process tends to be longer than  for \gls{CPU} and \gls{GPU}s, but shorter than the design of \gls{ASICs}. To alleviate this issue, \gls{FPGAs} vendors have introduced several \gls{CAD} tools that aim to facilitate the design process. For the particular context of \gls{AI} systems design based on \gls{FPGAs}, various efforts have been carried out; the reader is directed to \cite{sota_venieris_2018} for a detailed description. In this work, we analyze several academic tool flows used to map \gls{CNN}s onto \gls{FPGAs} accelerators. On the other hand, there are different tools that have been developed by \gls{FPGAs} vendors, which will be discussed as well.

\begin{figure*}[h]
\centering
\includegraphics[width=0.6\linewidth]{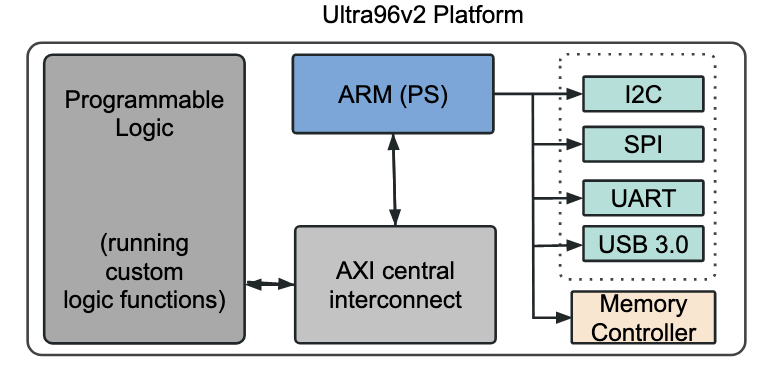}
\caption{Base hardware architecture used for this project. Images are fed through a USB 3.0 and then an \gls{MPSoC} device (Ultra96v2) is in charge of processing the video stream by using the integrated Application Processing Unit (or PS, processing system) to run the Operative System (OS), whereas the Programmable Logic section of the \gls{SoC} implements a neural network accelerator (custom logic).}
\label{img:SystemArch}
\end{figure*}

Considering all the features of \gls{FPGAs} discussed above, we posit that \gls{FPGAs} are excellent candidates for implementing smart cameras, as they do provide a good amount of computing capabilities while maintaining a low power consumption for specific tasks at the edge. Thus, in the work presented herein we discuss the implementation of a stand-alone smart camera capable of performing image segmentation of jet flames by performing most of the processing in an accelerator implemented in an \gls{FPGAs}, as shown in Figure \ref{img:SystemArch}. At the core of such smart platform there are \gls{AI}-models that are accelerated using the reconfigurable logic of the \gls{SoC} \gls{FPGAs}. However, directly implementing such models is not straightforward, as we will discuss next. 

\subsection{Neural networks on smart cameras}
{
Nowadays, it is clear that \gls{CNN} models can provide high accuracy results on different image processing tasks \citep{sota_vestias_2019,sota_venieris_2018} and many applications could benefit if implemented on edge devices. Unfortunately, edge devices like smart cameras are expected to be limited on hardware resources and power requirements and by contrast, most \gls{NN} implementations require a high amount of computational resources for execution and storing model parameters, leading to a high power consumption \citep{sota_venieris_2018,sota_vestias_2019,sota_berthelier_2021}. These issues make it then difficult to effectively exploit the capabilities of recently developed \gls{DL} models into edge devices with very limited computation power, in comparison to data center \gls{GPU}s used during training of such \gls{DL} models. In order to overcome these limitations, several researchers have advocated for the use of optimized \gls{DL} models whose numerical representation can be efficiently executed by less capable hardware, as it is the case for edge devices. According to \cite{sota_berthelier_2021} there are two main groups of techniques to optimize \gls{DL} models. The first group includes different compression techniques which aim to reduce the size of a model either by taking advantage of the redundancy of the \gls{DL} models (pruning and hashing) or by changing the numerical representation of the \gls{DL} model (quantization and binarization). A second group of techniques seeks to develop optimized architectures either by using more efficient layer arrangements or by searching more efficient topologies with techniques such as neuroevolution and supergraphs \citep{sota_berthelier_2021}. Some \gls{DL} frameworks, such as Tensorflow \citep{sota_tensorflow_2015} or Pytorch \citep{sota_pytorch_2019}, have developed utilities to compress \gls{DL} models by using techniques such as quantization and pruning. The resulting modules can be loaded in smaller devices such as \gls{CPU}s and even 8-bit micro-controllers.

In addition to these academic efforts, hardware vendors have also implemented their own tool-chains that are capable of porting and deploying \gls{DL} models to their specific hardware platforms by using either quantization and/or pruning techniques. In Table \ref{tab: sota_platforms_t} we summarize different software tools developed by hardware vendors. On this front, ARM has developed ARM \gls{NN} and CMSIS \gls{NN} to target their Cortex-A and Cortex-M microprocessors, Mali \gls{GPU}s, and more recently, their \gls{NPU}. At the same time, NVIDIA has created the JetPack\texttrademark\ software stack to allow their Jetson\texttrademark\ platform to run and accelerate different \gls{AI} applications. In a similar fashion, Intel has created the OpenVINO\texttrademark\ toolkit, which targets a wide number of devices such as \gls{CPU}s and \gls{GPU}s. Moreover, Intel has also developed the Intel FPGA AI Suite, which aims to accelerate inference by using accelerators implemented on \gls{FPGAs}. Finally, Xilinx has created Vitis AI for porting neural network models to their customizable \gls{FPGAs}-based accelerators which can target low cost and low power devices.

\gls{SoC} \gls{FPGAs} represent an ideal choice for implementing smart cameras as they integrate highly specialized \gls{PS} with several peripherals for sensor and video processing, as depicted in Fig. \ref{img:SystemArch}. The \gls{PS} can also be used to run specialized \gls{OS}s and to provide on-the-fly reconfiguration of the programmable component of the \gls{SoC} \gls{FPGAs}. The capability of integrating custom accelerators such as image processing cores and \gls{AI} model blocks  seamlessly is a promising area of research. Thus, we decided to use a Xilinx \gls{MPSoC} device to accelerate the inference of a semantic segmentation model. For this research, we considered two different Xilinx frameworks: \gls{FINN}, which is based on the works from \cite{sota_Umuroglu_2017} and \cite{sota_blott_2018}, and Vitis AI. Despite the fact that both frameworks provide an end-to-end tool-chain that enables the implementation and deployment of \gls{FPGAs}-based accelerators for \gls{NN}s, they use different strategies which have an impact on aspects such as cost, parallelism, latency, and power consumption \citep{sota_venieris_2018}. On the one hand, \gls{FINN} creates architectures of the streaming type which map each layer to an \gls{FPGAs} compute engine. This implies that the bigger the \gls{DL} model, the more hardware resources the accelerator will require. Thus, this type of accelerator is suitable for models which are to be run on large \gls{FPGAs} \citep{sota_venieris_2018}. On the other hand, single computing engine architectures are formed by at least one \gls{PE} which is in charge of performing different operations such as matrix multiplication and pooling. With this approach, every operation is mapped to a hardware-accelerated micro instruction which means that a set of consecutive micro instructions can represent all layers of a \gls{DL} model. Finally, a scheduler is in charge of controlling the hardware calls and dataflow \citep{sota_venieris_2018}. This approach is used by Vitis AI and it allows one to customize the accelerator depending on the \gls{FPGAs} resources. Since we used a small \gls{FPGAs} device, we opted for Vitis AI (version 1.3.2).

\subsection{Smart cameras on the edge}

There have been previous works oriented towards the implementation of smart cameras on the edge for specific applications. For instance, \cite{sota_giordano_2020} implemented a battery-free smart camera for face classification. The system can classify up to 5 different faces and then send the result through a \gls{LoRa} link which can cover several kilometers. To achieve such a goal, the ultra-low power system relies on a small \gls{NN} model of 39,821 parameters and 8-bit quantization to speed up the inference; also, a solar panel is used to harvest energy which is stored in a capacitor. When the capacitor is fully charged, the system takes a picture and tries to classify it; then the response is sent over a \gls{LoRa} link. One important factor for this system to work is the size of the \gls{NN} model, since it allows the use of a low-power consumption ARM Cortex-M4F microcontroller to run an \gls{AI}-based application. 

Interest in smart devices increased significantly during the COVID pandemic. For instance, an \gls{AI} network-based smart camera system prototype, which tracked social distance using a bird’s-eye perspective, was developed \citep{sota_up_25a}. Several \gls{CNN}s and object detection models were tested to identify people in video sequences. The final prototype, based on a Faster R-CNN model, was integrated in an embedded system. The software was developed using the NVIDIA Jetson Nano development kit and a Raspberry Pi camera module.

In the industrial sector (e.g., oil and gas, construction) it has also been recognized that big data analytics, powered by \gls{IoT} devices and smart cameras could be a game changer \citep{sota_up_25b}. In particular, several research works have suggested that the adoption of computer vision technology to enhance safety measures within manufacturing facilities and construction, focusing on workers and infrastructure, is of paramount importance.

An example in the food industry is presented in \cite{sota_guo_2020}, where an accessible vision system for small and midsize facilities is introduced. In this work, the authors propose to speed up the assessment of the quality of different food products through a visual inspection algorithm, which is capable of running at a rate of 100 \gls{FPS}. The authors used genetic algorithms to find useful features that can then be used by an AdaBoost algorithm in the classifier. They also claimed that this approach was more efficient than using a regular \gls{NN} approach. As a processing unit, they used the NVIDIA TX1 module which includes a \gls{GPU} and a Quad-Core ARM cortex \gls{CPU}.

A thorough study presented by \cite{sota_up_25d} identifies and analyzes how computer vision and smart cameras could enhance safety management across various sectors within the Safety 4.0 paradigm. Several key safety factors are discussed, such as accident prevention, risk management and real-time monitoring, across sectors, including construction \citep{sota_up_25e}, industrial safety, disaster and public safety \citep{sota_up_25f}. The analysis reveals a significant trend towards shifting from reactive to proactive safety management, facilitated by the convergence of \gls{AI} with \gls{IoT} and Big Data analytics.

For instance, the research by \cite{sota_up_25c} introduces SafeFac, an intelligent camera-based system for managing the safety of factory environments. In their approach, several cameras are integrated on the assembly line to detect instances of workers approaching machinery in hazardous situations and to alert system managers and stop the line if needed. Given the complex setting (multi-camera and low response latency), the proposed approach exploits a YOLOv3-based light-weight human object detection. To address the latency–accuracy trade-off, SafeFac incorporates a set of algorithms as pre- and post-processing modules, and a novel adaptive camera scheduling scheme.

Some more recent works have started to focus on the hazards associated with fire and industrial safety. In most industrial settings, fire safety inspectors are responsible for visual inspection of safety equipment and reporting defects. As traditional approaches involve manually checking each piece of equipment, which is time-consuming and inaccurate, several works have been proposed to improve the inspection process \citep{sota_up_25g}. Using computer vision and \gls{DL} techniques, object detection models have been trained to visually inspect fire extinguishers and identify defects. However, such work does not address smart camera or mobile (i.e., smart-phone based) implementations of the proposed models. The same issue applies for a recent work in fire inspection equipment \citep{sota_up_25i}. This paper presents a Line of Fire (LoF) Detection system that leverages real-time video feeds, object detection, and motion tracking to identify and alert when individuals are in the path pipes.  This study presents a system to automate fire safety inspection tasks, including detecting fire safety equipment (FSE), but no mobile or smart-camera implementation is presented.

Finally, a recent work more aligned to our research studies vision-based patterns of fire events to identify multiple objects that contribute to different types of fire accidents. To achieve this goal, a \gls{CNN} based on \gls{DL} is applied \citep{sota_up_25j}. Flames and smoke are trained as multiple objects to recognize fire event patterns, while their size and position are visualized to assess fire severity.

In summary, smart cameras have a wide range of applications; however, there are two main critical situations which limit their usability: power consumption and processing frame rate. The reviewed literature shows that there is a wide range of applications for smart cameras, and their potential is clear for many of such applications. Various prototypes of such smart cameras have been created, and most of them try to improve the performance in different aspects: frame rate, accuracy, and power consumption.

\begin{figure*}[t]
\centering
\includegraphics[width=0.9\linewidth]{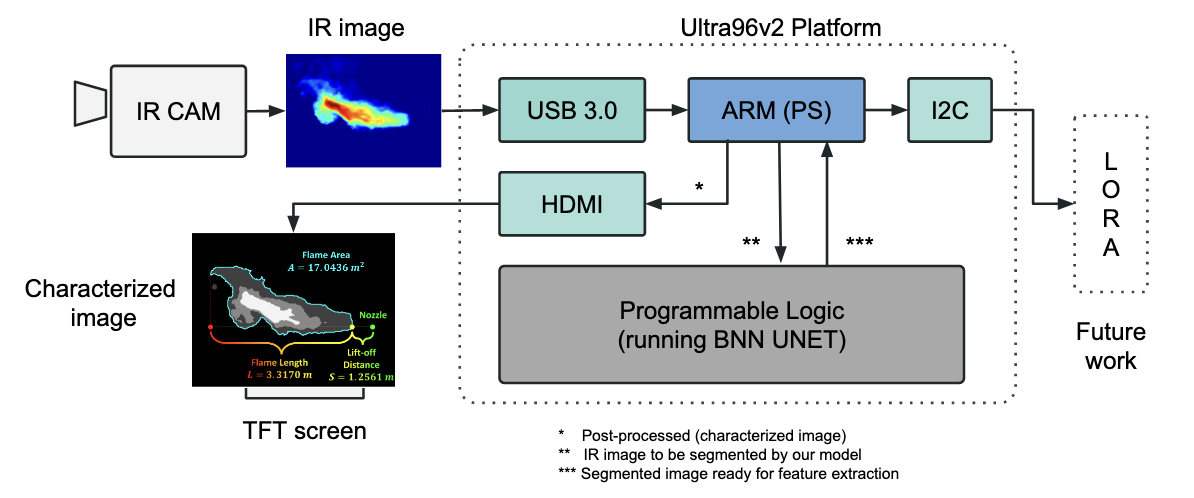}
\caption{Proposed solution model for implementing a smart camera for jet fire risk assessment. The Ultra96 board is connected to an IR camera via an USB peripheral within the SoC. The input image is processed by the PS block (an ARM processor) and fed to the PL section where our binarized UNet model resides. After processing the image, this is fed back to the PS to perform feature extraction. These features can be used for visualization purposes on a TFT screen (in the lab) or can be sent  via an communication protocol (i.e., LORA) to a cloud provider to implement a full-fledged IIoT solution. Herein, we haven't implemented these functionalities yet and is left for future work. All the communication between the PS and the PL and peripheral is done through an AXI interconnect, not shown here in order to simplify the diagram.}
\label{img:SystemFinal}
\end{figure*}

\begin{table*}
\caption{Mapping tool-chains developed by hardware vendors to deploy neural networks on their devices.}
\label{tab: sota_platforms_t}
\setlength{\tabcolsep}{3pt}
\begin{tabular}{|p{90pt}|p{60pt}|p{190pt}|}
\hline
\textbf{Tool-chain}& 
\textbf{Developer}& 
\textbf{Devices} \\
\hline
ARM NN & ARM & ARM-Cortex-A CPU, Mali GPU, Ethos Neural Processor Unit (NPU) \\
CMSIS NN & ARM & ARM-Cortex-A and ARM-Cortex-M CPUs \\
Vitis AI & XILINX (AMD) & Deep Learning Processor Unit (DPU) implemented in a Field Programmable Gate Array (FPGAs), Multiprocessing SOC (MPSOC), System on module (SOM) \\
FINN & XILINX (AMD) & It's an experimental framework focused on quantized neural networks, which creates custom architectures for each model\\
OpenVino & Intel & Intel Core, Atom and Xeon CPUs, Integrated GPUs, Vision Processing Units (VPUs) and, FPGAs \\
Intel FPGA AI Suite& Intel & Agilex, Cyclone 10 GX and, Arria 10 FPGAs \\
JetPack & Nvidia & Different Jetson development kits \\
\hline
\multicolumn{3}{p{236pt}}{}\\
\end{tabular}
\end{table*}

\section{Methodology}
\label{Methodology}


In this paper we propose a custom smart camera application for \gls{IR} flame segmentation based on an \gls{FPGAs} accelerator by using optimized hardware and optimized algorithms.

The system we have implemented is depicted schematically in Fig. \ref{img:SystemFinal} and it is intended for experimental and proof-of-concept purposes. The main goal of the platform is to implement image-processing tasks (using an onboard \gls{CNN}), as well as post-processing tasks for geometric jet fire characterization in real-time.

The system works as follows: first, it processes \gls{IR} images captured from an infrared camera via USB 3.0; then, the processing components (\gls{PS}, an \gls{ARM} processor) manage the transfer of the input frames to the \gls{PL}, which runs a segmentation model based on a UNet. The UNet herein is a model optimized for efficient hardware acceleration.  The segmentation masks generated by the module are subsequently used for flame characterization back in the \gls{PS}. Finally, the original segmentation and the measurements are displayed on a screen. Several peripherals of the \gls{SoC} \gls{FPGAs} are exploited (\gls{OS}, drivers, functions) for systems management, which run on the \gls{PS} component of the chip.

In summary, the developed platform is used to create a laboratory proof of concept whose goal is to serve as an experimental setup.

\subsection{Case study}
\label{Case study}

\gls{IR} videos obtained with an Optris PI 640i camera (spectral range: 8–14 $\mu$m) of horizontal propane jet fires captured in a laboratory setup (see \citet{method_foroughi_2021}) were used in this work. The \gls{IR} camera was running at 4 \gls{FPS}, resulting in a total of 201 images with a resolution of 640 x 480 pixels.

The geometric characteristics of jet flames can be determined through image segmentation techniques applied to \gls{IR} thermal imaging data. As proposed by \citet{method_foroughi_2021} and represented in Figure \ref{img:firecharacteristics}(c), the flame structure can be partitioned into three distinct temperature regions: i) a high-temperature central zone representing the flame core, ii) an intermediate-temperature middle zone corresponding to the reaction zone, and iii) a lower-temperature outer zone encompassing the flame periphery and mixing regions. The segmentation of these temperature zones was carried} out in this case using automated image processing segmentation models and validated by experts in the field, more details can be found in \cite{carmina_journal}.

Following segmentation, geometric parameters, including flame area, flame length, and lift-off distance, were quantified using the masked temperature zone representations in conjunction with metric-per-pixel scaling ratios and precise exit nozzle pixel coordinates. The flame area was calculated by determining the total pixel count within the segmented contours, while flame length and lift-off distance were derived from the dimensional properties of the contour bounding boxes.

The inherent dynamic nature of turbulent jet flames presents significant challenges for conventional temperature-based thresholding approaches aiming to achieve robust and consistent segmentation results across varying flame conditions and temporal fluctuations. Therefore, this semi-automatic segmentation and validation was essential.

\begin{table}
\caption{DPU configuration.}
\label{tab: methodology_dpu_t}
\setlength{\tabcolsep}{3pt}
\begin{tabular}{|p{130pt}|p{90pt}|}
\hline
\textbf{Feature} & \textbf{Supported}\\
\hline
Clock Frequency & 300MHz\\
Double Data Rate & Yes \\
Throughput & Up to 2304 per clock cycle \\
Channel Augmentation & Yes \\
Convolution Layers & Yes \\
Depthwise Convolution Layers & Yes \\
Max pool Layer & Yes \\
ReLu and Leaky ReLu activation & Yes \\
\hline
\multicolumn{2}{p{200pt}}{}\\
\end{tabular}
\end{table}

\subsection{Vitis AI software stack}

\begin{figure*}[t]
\centering
\includegraphics[width=0.5\linewidth]{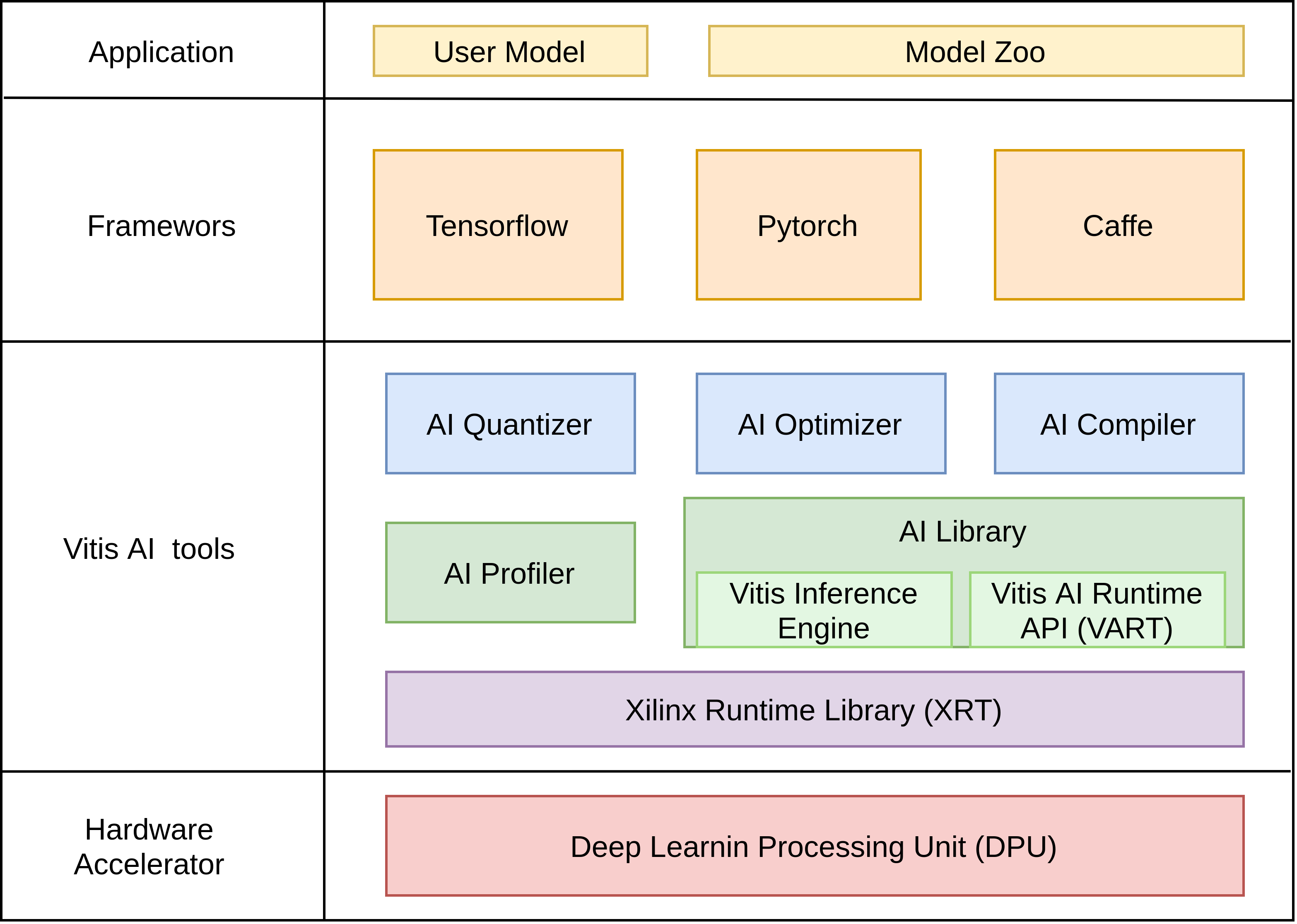}
\caption{Software stack of Vitis \gls{AI}.}
\label{img:VitisAi}
\end{figure*}

As described in Table \ref{tab: sota_platforms_t}, the Vitis \gls{AI} is a framework developed by Xilinx that aims to facilitate the portability of \gls{AI} models to their hardware platforms. Beyond the model optimization tools, the Vitis \gls{AI} suite provides an \gls{API} for both Python and C++ which speeds up the development of a complete application focusing on reducing prototyping time. Fig. \ref{img:VitisAi} shows the software stack of this suite.

\begin{figure*}[t]
\centering
\includegraphics[width=0.55\linewidth]{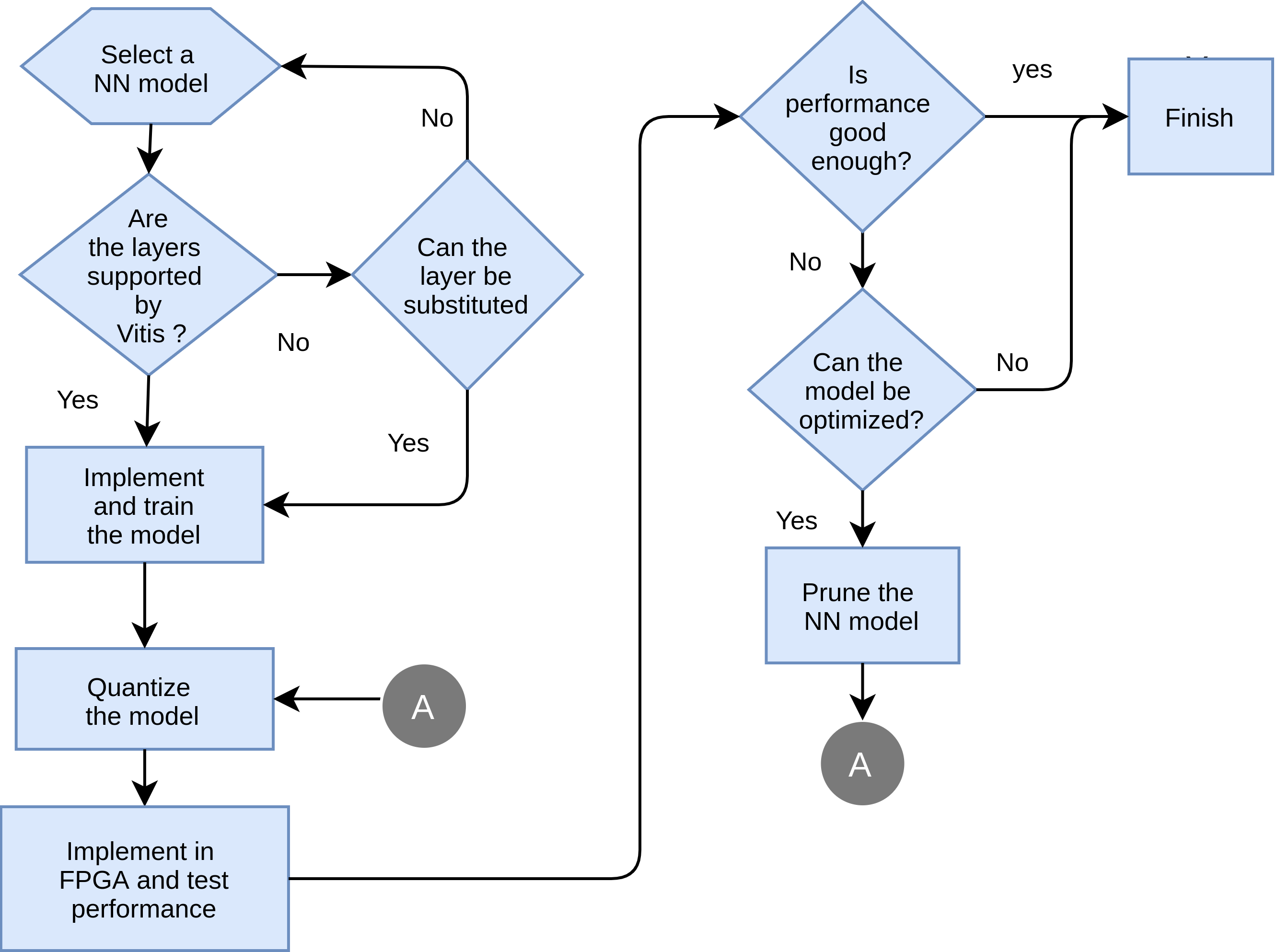}
\caption{Process used to port a neural network to an embedded system extending from Vitis \gls{AI} flow.}
\label{img:flowchart}
\end{figure*}

It is worth noting that not all operations can be accelerated by the \gls{DPU}. Therefore, if a specific layer architecture is not supported, a \gls{CPU} implementation can still be provided, but this solution may have a negative impact in terms of performance. For this work, only Vitis-supported layers were used. Fig. \ref{img:flowchart} shows the algorithm used in this work to ensure that all operations required by the \gls{DL} model were supported by the \gls{DPU}.

\subsection{Full precision model}
Classifying the temperature of a jet flame region based on its infrared spectrum can be approached as an image segmentation problem, as demonstrated by \cite{motiv_perez_2021}. Consequently, a UNet model  topology \citep{method_ronneberger_2015} was selected as the baseline . To ensure the final network would fit within the \gls{DPU}, the number of filters in the deepest layers was reduced. Two different models were trained: one with only the number of filters reduced, and the other with both filter reduction and batch normalization layers added between each convolution layer. Before applying any optimization, such as pruning, both models were trained. The first model had 7,481,028 trainable parameters, while the second model, which included \gls{BN}, had 6,151,748 trainable parameters. For comparison, the original UNet model contains 31,024,832 trainable parameters \citep{method_beheshti_2020}, representing reductions of 75.89\% and 80.17\%, respectively. This substantial decrease in trainable parameters translates into faster training and inference times. Furthermore, research such as Squeeze UNet \citep{method_beheshti_2020} has explored alternative strategies to further reduce the number of filters, demonstrating significant improvements in latency, memory usage, the number of \gls{MACs} and, consequently, energy efficiency.

Our model was trained with the dataset of \gls{IR} images presented in section \ref{Case study}. Thus, in order to speed up the inference process and boost the attainable frame rate, we downscaled these images by a factor of 2 in both the x and y axes. The dataset was then split according to the following distribution: 96 images for training, 50 for validation during training, and 55 images for testing. To compensate for the small number of training images we used the \textit{Albumentations} tool \citep{method_Buslaev_2018} to create a data augmentation pipeline conformed by geometric transformations (i.e., rotation, flips and shifts). More details are given in Table \ref{tab: methodology_dataaug_t}.

\begin{table}
\caption{Data augmentation pipeline.}
\label{tab: methodology_dataaug_t}
\setlength{\tabcolsep}{3pt}
\begin{tabular}{|p{70pt}|p{60pt}|p{80pt}|}
\hline
\textbf{Operation} & \textbf{Probability} & \textbf{Range of Values}\\
\hline
Vertical Flip & 0.5 & \\
Horizontal Flip & 0.5 & \\
Horizontal shift & 0.66 & $\pm$ 10\% of the pixels\\
Vertical shift & 0.66 & $\pm$ 10\% of the pixels\\
Rotation & 0.66 & $\pm$ 30 degrees\\
\hline
\multicolumn{3}{p{200pt}}{}\\
\end{tabular}
\end{table}

One challenge that occurred while training the model with the jet flame \gls{IR} images, was that they were composed of a small area of flame over a significantly larger background, which generated an imbalance in the pixels of the image. In order to compensate for such imbalance, the chosen loss function was Weighted Categorical Cross Entropy. The weights used in the loss function were calculated according to Eq. \ref{eq:weights}, a formula proposed by \cite{method_paszke_2016}. 

\begin{equation}
\label{eq:weights}
    W_{class} = \frac{1} {ln(1.02+P_{class})}
\end{equation}

Where \(W_{class}\) and \(P_{class}\) represent the class weight and class probabilities

The final values for the weights were as follows: 1.5901, 10.5801, 17.2133, and 22.4526 (for the three areas as well as the background). The selected optimizer was \gls{SGDW} \citep{method_loshchilov_2017}. As a form of adding regularization during the training process, \gls{SGDW} was configured to use a Nesterov momentum of 0.98 (after several trial and error iterations). 

Both learning and weight decay rates were defined as a cyclical polynomial decay function with a power of 0.5 over 45000 training steps. The initial value for the learning rate was 0.01 and the final value was 0.005, while the initial value for weight decay rate was 0.0005 and the final value was 0.00025. 

The evaluation metric during training was set to the Dice coefficient. This metric was proposed by \cite{method_dice_1945} and \cite{method_sorensen_1948}, and aims to measure the similarity of two different sets. When the coefficient is 1.0, it means that the sets are equal. During the training process WCE was used as loss function, and the the Dice metric was only used for scheduling the lerning rate and defining the early stopping criteria.

To improve the performance of the floating point model, in addition to all previous settings used, we also constrained every convolutional layer filter to have a restricted norm in the range [0, 1.0] . This measure is used to decrease the dynamic range of the values learned by the filters (along with weight decay). The bias of every filter was also constrained to a maximum norm of 6.0. The rationale for this choice was that by decreasing the dynamic range of the values that a filter can have, the decrease in precision by the time the model is quantized to 8-bit values will have less impact. 

As an alternative, a second model was trained with the same previous settings replacing the original evaluation metric (i.e., Dice coefficient) by the Jaccard index. The Dice index (also called Dice coefficient, Dice score, or Sørensen–Dice coefficient) is a measure of similarity between two sets . It calculates the area of overlap between the predicted segmentation and the ground truth. The Dice index handles class imbalance better than accuracy, and works well with small structures and directly measures overlap between masks, making suitable for our purposed.  Yhe closer it is to a value of 1, the better the performance. The index was also used as a means for early stop criteria. 

For the second set of experiments integrating \gls{BN}, the chosen loss function was  Cross Entropy, widely used in classification tasks.  It quantifies the difference between predicted probability distributions and true class labels. This function penalizes incorrect, overconfident predictions and promotes more accurate, well-calibrated estimates. As a result, it helps train models to distinguish between multiple classes by maximizing the likelihood of the correct class. In our experimets for this configuration, cross entropy yielded better results than Weighted Categorical Cross Entropy, and thus was retained. The selected optimizer was Adam \citep{Kingma2017}, with a learning rate of 0.0001, a batch size of 20, and 350 training epochs. \gls{BN} \citep{Ioffe2015} was used to stabilize and accelerate learning by normalizing each layer’s inputs. This approach minimizes internal covariate shift by maintaining consistent data distributions, improving gradient flow. \gls{BN} also introduces mild regularization and reduces reliance on other techniques. Together, these strategies allow faster convergence and often improve deep neural network performance.

\subsection{\gls{DL} model optimization}

\subsubsection{Quantization of the network}
After training the model, we proceeded to quantize the full-fledged architecture. We performed this step using the Vitis AI quantizer module which was configured to apply \gls{CLE} \citep{method_nagel_2019}, a technique that attempts to improve the calibration performance during quantization. \gls{CLE} was applied during 30 epochs. The result was a \gls{CNN} model with all values represented with only 8 bits. After confirming that there was no significant degradation in the model performance (in terms of segmentation accuracy), we proceeded to compile the quantized model with the Vitis compiler which created an \textit{xmodel} file with all the instructions required by the \gls{DPU} to execute the model.

\subsubsection{Pruning the network}
A basic application was developed using PYNQ \citep{method_pynq}, an easy-to-use framework developed by Xilinx that enables speeding up the \gls{FPGAs} prototyping task by providing a \gls{HAL}. Herein, we used the python based \gls{VART} \gls{API} to control the instantiated \gls{DPU}. This implementation yielded a segmentation performance comparable to that of the floating point UNet model used in \cite{motiv_perez_2021}. Despite of this good accuracy, the attainable frame rate was far from optimal or not as good as we expected for meeting our edge computing requirements for this initial configuration. Thus, we further optimized our UNet model by using coarse pruning techniques. For this purpose, we made use of \textit{Keras Surgeon} \citep{method_whetton_2017}, a tool designed to work with keras/tensorflow models. The advantage of using coarse pruning is that some filters are completely removed from the model and this decreases the amount of calculations required by the model. This is important since the main bottleneck of \gls{CNN}s running on constrained devices is the number of computations required \citep{sota_vestias_2019}, which is directly proportional to the number of filters. Thus, the next step was to identify the filters to be removed, a process that entails the calculation of the \gls{APoZ} \citep{method_hu_2016}, which helps to identify how much a neuron contributes to the model. After identifying the \gls{APoZ} for all the neurons, we removed 3\% of the filters with the highest \gls{APoZ}. Afterward, the new model was fine-tuned to recover from the loss of neurons. 

After this process, the model was evaluated and its performance was compared against the original optimized implementation of the model; this process was repeated until the model loss did not longer decrease (around 3\%). This threshold was selected after several experiments that showed that setting a smaller threshold (1\%) would remove only a small fraction of filters, increasing the pruning time significantly. On the other hand, setting the threshold to a higher value (5\%) would improve the filter elimination and reduce the pruning time but it would significantly decrease the performance of the model and its ability to recover the original performance. It is important to note that the 2 first and 2 last convolution layers were not pruned since this would cause a big drop in the performance of the model.

\subsection{Hardware implementation}

For this project, we chose the Ultra96V2 development board from Avnet, a low-cost platform that is supported by the Vitis \gls{AI} tool flow. The board includes a ZU3EG A484 chip that belongs to the Xilinx Zynq UltraScale+Mpsoc family. The \gls{SoC} is mainly encompassed internally by a \gls{PS} and a \gls{PL} unit. The \gls{PL} can be used to implement different hardware accelerators which can offload \gls{PS} logic from compute-intensive tasks. In this work, we used the \gls{PL} logic to create a single instance of the \gls{DPU}, an \gls{IP} developed by Xilinx which speeds up the inference of \gls{CNN}.

From an architectural point of view, the \gls{DPU} fits in the category of a single computation engine accelerators described in \cite{sota_venieris_2018}. This is important since such types of accelerators can be customized to specific resource availabilities and hence they are a better fit for devices with limited resources. 

The system implementation of our solution is shown in Figure \ref{img:SystemFinal}. The \gls{PS} is in charge of coordinating the entire life-cycle of our application: retrieving images acquired from the camera, feeding them to the \gls{PL} section of the \gls{SoC} (hardware accelerator implementing the UNet pruned model) and processing the segmented image to extract the features for characterizing the jet fire. 

For the hardware implementation, XRT, Vitis, and Vitis \gls{AI} were used to obtain the .bit files, which contain the configuration data for the \gls{FPGAs}. Table \ref{tab: methodology_dpu_t} shows the \gls{DPU} configuration used for the implementations presented in this paper. Those values were obtained experimentally to maximize the achievable throughput (maximum FPS). 

\begin{figure*}[t]
\centering
\includegraphics[width=0.65\linewidth]{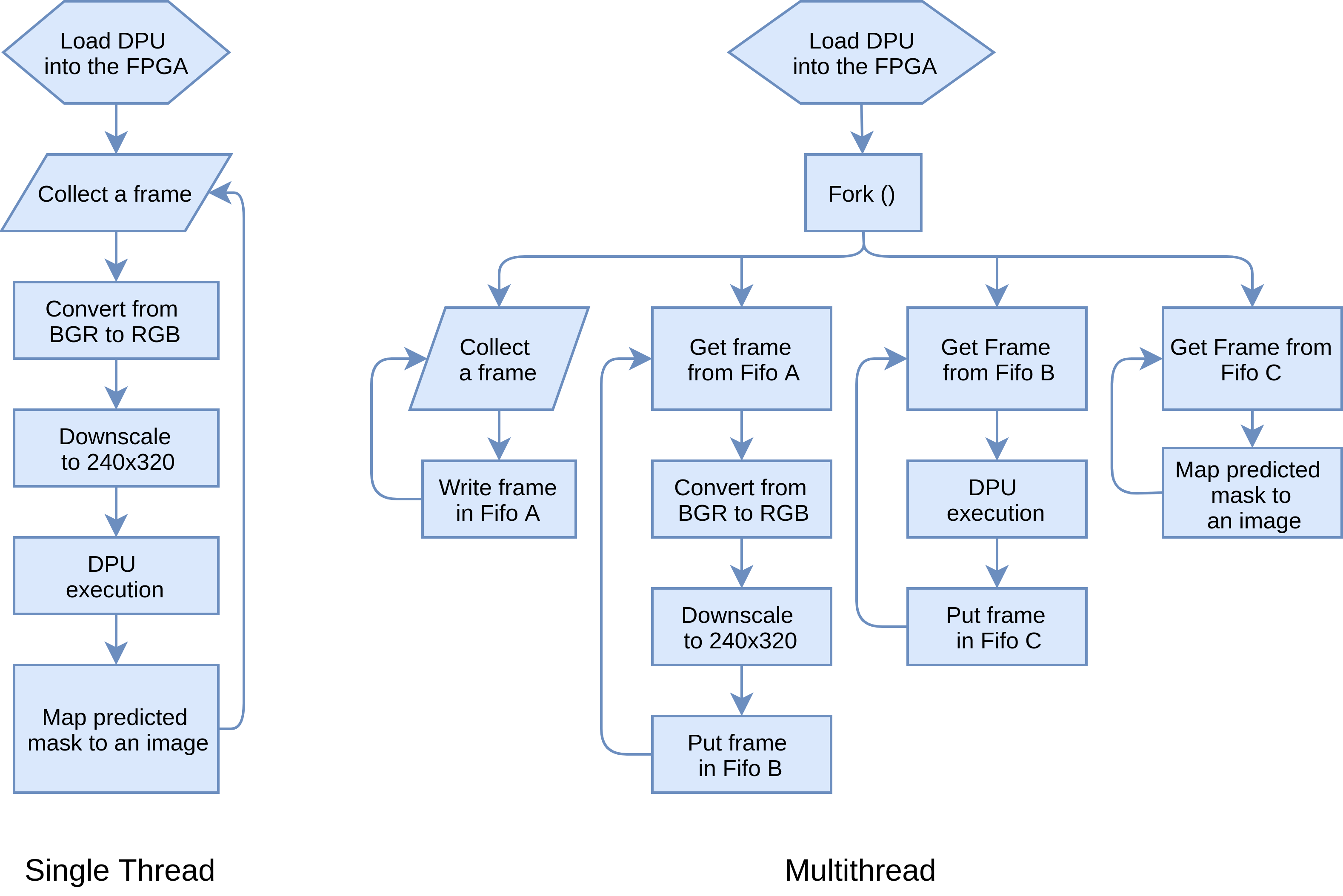}
\caption{Flow chart for single- and multi-threading inference approaches.}
\label{img:SingleAndMultiThread}
\end{figure*}

\subsection{Software implementation}

After developing the initial software prototype, we found that overall performance was insufficient, even for the pruned models. A detailed latency analysis of each pipeline stage revealed that the primary bottleneck was the \gls{DPU} during the inference phase. To address this issue, we implemented a multi-threaded approach, enabling better performance by overlapping processes. Thread communication was managed using queues.


Fig. \ref{img:SingleAndMultiThread} shows the flow charts for both software implementations. It illustrates the inference approaches used by the \gls{PL} to process images captured by the camera. The implementation utilizes Python for preprocessing, postprocessing, loading the model onto the \gls{DPU} with the .bit file (for \gls{DPU} configuration), the .xmodel (for the compiled model), and managing communication with the \gls{DPU} to perform inference with the UNet model for fire segmentation. In the single-threaded approach, these steps occur sequentially. In contrast, the multi-threaded approach executes these tasks in parallel, optimizing the time required to process each camera frame. Communication between processes is managed using queues.

The execution process works as follows:
\begin{enumerate}
    \item In the first process, the frame is collected by the \gls{PL}.
    \item In the second process, the image is preprocessed before inference, the image is converted from BGR to RGB, and downscaled to \(240\times320\).
    \item In the third process, there is \gls{DPU} execution, in which the preprocessed frame is passed through the UNet model for fire segmentation. 
    \item In the fourth process, the prediction is postprocessed into an image.
\end{enumerate}


\subsection{Computational performance evaluation}

Another aspect of utmost importance is that the model can attain an acceptable performance in real-time settings, being 24 \gls{FPS} the lower bound. For this, evaluating the computational complexity in terms of \gls{MACs} and \gls{FLOPs}, as well as the inference time, is essential. 

The calculation of the number of \gls{MACs} and \gls{FLOPs} was done following the equations defined by \cite{Molchanov2017}, \cite{Getzner2023} and \cite{Press2007} for the different types of operation (i.e., convolution, maximum pooling layers, ReLU, upsampling layers, \gls{BN} layers):

\newpage

\begin{itemize}

\item \textbf{Multiply-Accumulate Operations (\gls{MACs}) for convolution}

 For calculating the number of \gls{MACs} for convolutional kernels we used Eq. \ref{eq:macs_conv}.

\begin{equation}
\label{eq:macs_conv}
    \text{MACs} = H \cdot W \cdot C_{in} \cdot C_{out} \cdot K^{2}
\end{equation}

Where \(H\), \(W\) and \(C_{in}\) are the height, width, and number of input channels of the feature map, \(K\) is the kernel width (assuming it is symmetric) and \(C_{out}\) is the number of output channels.

\item \textbf{Floating-Point Operations (\gls{FLOPs}) for convolution}

The number of \gls{FLOPs} for convolution was obtained according to Eq. \ref{eq:flops_conv}. 

\begin{equation}
\label{eq:flops_conv}
    \text{FLOPs} = 2 \cdot H \cdot W \cdot (C_{in} \cdot K^{2} + 1) \cdot  C_{out}
\end{equation}

\item \textbf{Multiply-Accumulate Operations (\gls{MACs}) for max pooling layers}

For calculating the number of \gls{MACs} of a max pooling layer we used Eq. \ref{eq:macs_max}. 

\begin{equation}
\label{eq:macs_max}
    \text{MACs} = (K^{2} \cdot W_{out} \cdot H_{out} \cdot C_{in})/2
\end{equation}

Where \(K\) is the kernel size of the maximum pooling window, \(W_{out}\) and \(H_{out}\) are the output weight and height, and \(C_{in}\) is the number of input channels.

Note that the maximum pooling layer does not perform any multiply-accumulate operations, so the \gls{MACs} are approximated by dividing the number of \gls{FLOPs} (see Eq. \ref{eq:flops_max}) by two.

\item \textbf{Floating-Point Operations (\gls{FLOPs}) for maximum pooling layers}

The number of \gls{FLOPs} was calculated according to Eq. \ref{eq:flops_max}.

\begin{equation}
\label{eq:flops_max}
    \text{FLOPs} = K^{2} \cdot W_{out} \cdot H_{out} \cdot C_{in}
\end{equation}

\item \textbf{Multiply-Accumulate Operations (\gls{MACs}) for ReLU}

For the ReLU activation function, we followed the formulation shown in Eq. \ref{eq:macs_max_ReLU}.

\begin{equation}
\label{eq:macs_max_ReLU}
    \text{MACs} = (W_{in} \cdot H_{in} \cdot C_{in})/2
\end{equation}

Where \(W_{in}\), \(H_{in}\) are the input weight and height, and \(C_{in}\) is the number of input channels.

Note that the ReLU does not perform any multiply-accumulate operations, so the \gls{MACs} are approximated by dividing the number of \gls{FLOPs} (see Eq. \ref{eq:flops_max_ReLU}) by two.

\item \textbf{Floating-Point Operations (\gls{FLOPs}) for ReLU}

The number of \gls{FLOPs} for the ReLU activation function was derived as shown in Eq. \ref{eq:flops_max_ReLU}.

\begin{equation}
\label{eq:flops_max_ReLU}
    \text{FLOPs} = W_{in} \cdot H_{in} \cdot C_{in}
\end{equation}

\item \textbf{Multiply-Accumulate Operations (\gls{MACs}) and\\ Floating-Point Operations (\gls{FLOPs}) for upsampling layers}

The upsampling algorithm used was nearest neighbor, so the number of \gls{MACs} and \gls{FLOPs} was zero in both cases. Each output pixel simply copies the value of the nearest input pixel, resulting in pure data movement.

\item \textbf{Multiply-Accumulate Operations (\gls{MACs}) and\\ Floating-Point Operations (\gls{FLOPs}) for batch normalization layers}

While the batch normalization layer performs operations during training, Vitis AI fuses this layer with the preceding convolution during model optimization. As a result, during inference, the \gls{MACs} and \gls{FLOPs} attributed to batch normalization were zero.

\end{itemize}

\section{Results}
\label{Results}

\subsection{Quantitative evaluation of the similarity performance}

In the work from \cite{carmina_journal} it was determined that the metric with the highest correlation to the experts’ ranking was the \gls{JI}, which is a dissimilarity measure mostly used when boundary delineation of a segmentation is important. The \gls{JI} metric measures the extent to which each point of a segmentation lies near some point of the ground truth and vice versa; so it can be used to determine the degree of resemblance between two images when superimposed on each other. A value closer to one  is preferable to maximize this metric.

The Dice Index metric was used to compare the performance of the full-precision and the optimized models. In what follows, we discuss the results obtained for the six different implemented models, namely: i) the original full precision models (UNet models trained with Dice and Jaccard indexes), ii) the optimized models (Pruned UNet models trained with Dice and Jaccard indexes), and iii) the UNet models obtained using \gls{BN} layers (with and without pruning) and trained with Cross-Entropy. 

After applying quantization and pruning techniques for fitting the model in the reconfigurable logic, a drop in performance is expected. However, considering the Dice metric and according to the numerical results summarized in Table \ref{tab: methodology_model_perf} for the testing set, the UNet model trained with Dice index attained a value of 0.950, whilst the optimized model (i.e., Pruned UNet Dice ) exhibited only a slight drop in performance (a Dice value of 0.942 or 1.05 \%). Despite this slight drop, the number of parameters of this model is still fairly large (224,118). The UNet model trained using the Jaccard index followed a similar trend, but exhibited a larger drop in performance. The Jaccard trained model already exhibits a drop of 3.47\% compared to the base model and pruning the model  produces an overall drop of 6.63\%, making these two models unsuitable for our intended application (the Dice index for the Pruned model with Jaccard loss is 0.887), despite the significant reduction of parameters (from 7'4811,028 to 234,081).

The models trained with Batch Normalization and Cross Entropy performed much better, even surpassing the baseline model performance by 2.53\% and 2.42\%, respectively. Furthermore, the pruned UNet with batch normalization only accounts for a fraction of the original parameters (a reduction in model size of 99\%). This improvement is explained by the introduction of batch normalization that keeps activations centered and with stable variance and for its capability of acting a regularizer.  As we will see in Section 4.3, this has a tremendous impact in the overall model performances achieved by our porposed smart camera implementation.

\begin{table*}
\caption{Performance drop observed in different models after quantization. Performance values refer to Dice Index (DI). The perceptual drops in the last column are given for the quantized models against the base model, UNet (Dice)}
\label{tab: methodology_model_perf}
\setlength{\tabcolsep}{3pt}
\begin{tabular}{|p{205pt}|p{50pt}|p{60pt}|p{70pt}|p{60pt}|}
\hline
\textbf{Model (Metric)} & 
\textbf{Parameters Count} & 
\textbf{DI - Floating Point Model} &
\textbf{DI - Quantized model} & 
\textbf{Performance drop}\\
\hline
UNet (Dice) & 7'481,028 & 0.953 & 0.950 & N/A  \\
Pruned UNet (Dice) & 224,118 & 0.940 & 0.942 & -1.05 \% \\
UNet (Jaccard) & 7'481,028 & 0.918 & 0.917 & -3.47 \% \\
Pruned UNet (Jaccard) & 234,081 & 0.893 & 0.887 & -6.63 \% \\
UNet Batch Normalization (Cross Entropy) & 6'151,748 & \textbf{0.975} & \textbf{0.974} & +2.53\%\\
Pruned UNet Batch Normalization (Cross Entropy) & \textbf{59,095} & 0.974 & 0.973 & +2.42\%\\
\hline
\multicolumn{5}{p{240pt}}{}\\
\end{tabular}
\end{table*}



\subsection{Qualitative evaluation of the similarity performance}

In Table \ref{tab: results_visual} we can see 4 images from the experimental dataset, their corresponding ground truth images, and the prediction made by the models developed in the present work. The UNet models with \gls{BN} layers, both standard and pruned, consistently deliver very accurate and precise segmentation results in concordance with the quantitative results and the ground truth (Table 5) despite the model compression and optimization. Furthermore, the qualitative results  showcase that all models capture the complex shapes of the flames, sharply delineating boundaries, and accurately representing internal heat intensity distributions. In this sense, the models using Jaccard index perform consistently and are typically the closest the to \gls{BN} models.

In general, the integration of \gls{BN} enables models to produce sharper, more detailed, and more accurate fire segmentation across a variety of flame shapes and intensities, while pruned models show a remarkable quality retention, suggesting that such methods can be applied without significant quality loss. 


\begin{table*}[htpb!]
\newcolumntype{V}{>{\centering\arraybackslash}p{90pt}}

\caption{Visual comparison of the results obtained with different models for four different images from the jet flame case study and the corresponding ground truth. It is important to note that we are interested to maintain a similar performance across the models.} 
\label{tab: results_visual}

\setlength{\paperwidth}{3pt}

\begin{tabular}{|p{80pt}|V|V|V|V|}

\hline

Image & 
Example 1 &
Example 2 &
Example 3 &
Example 4 \\
\hline

Original Image &
\includegraphics[width=0.3\textwidth, height=24mm, keepaspectratio]{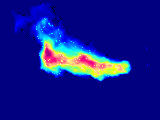} &
\includegraphics[width=0.3\textwidth, height=24mm, keepaspectratio]{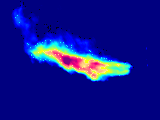} &
\includegraphics[width=0.3\textwidth, height=24mm, keepaspectratio]{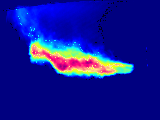} &

\includegraphics[width=0.3\textwidth, height=24mm, keepaspectratio]{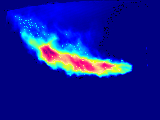} \\
\hline

Ground Truth & 
\includegraphics[width=0.3\textwidth, height=24mm, keepaspectratio,valign=c]{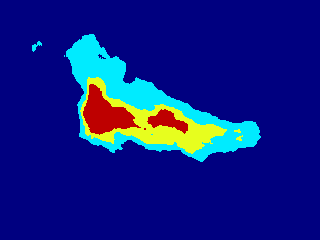} &
\includegraphics[width=0.3\textwidth, height=24mm, keepaspectratio,valign=c]{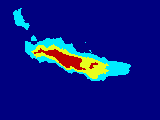} &
\includegraphics[width=0.3\textwidth, height=24mm, keepaspectratio,valign=c]{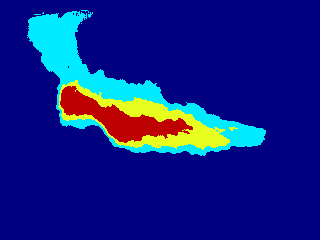} &
\includegraphics[width=0.3\textwidth, height=24mm, keepaspectratio,valign=c]{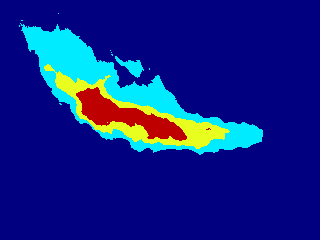} \\
\hline
 UNet (Dice) &
\includegraphics[width=0.3\textwidth, height=24mm, keepaspectratio,valign=c]{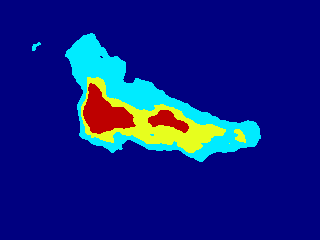} &
\includegraphics[width=0.3\textwidth, height=24mm, keepaspectratio,valign=c]{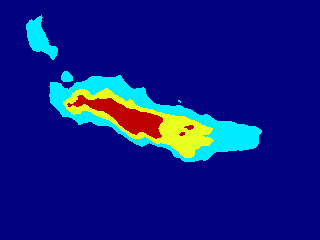} &
\includegraphics[width=0.3\textwidth, height=24mm, keepaspectratio,valign=c]{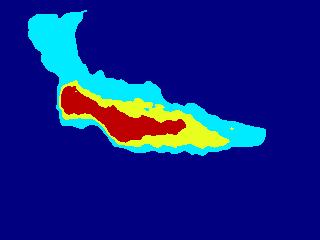} &
\includegraphics[width=0.3\textwidth, height=24mm, keepaspectratio,valign=c]{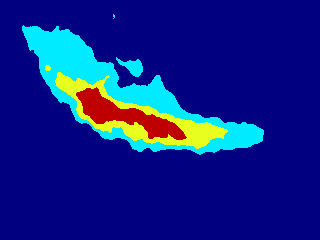} \\
\hline

Pruned UNet (Dice) &
\includegraphics[width=0.3\textwidth, height=24mm, keepaspectratio,valign=c]{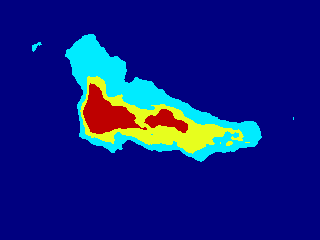} &
\includegraphics[width=0.3\textwidth, height=24mm, keepaspectratio,valign=c]{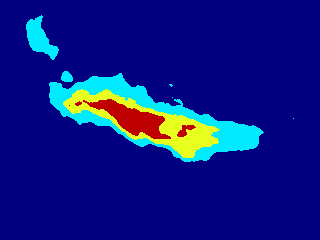} &
\includegraphics[width=0.3\textwidth, height=24mm, keepaspectratio,valign=c]{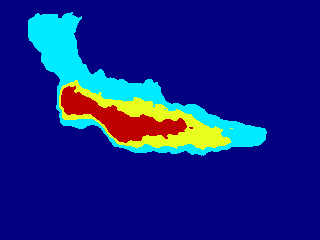} &
\includegraphics[width=0.3\textwidth, height=24mm, keepaspectratio,valign=c]{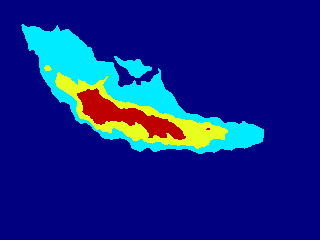} \\
\hline

UNet (Jaccard) &
\includegraphics[width=0.3\textwidth, height=24mm, keepaspectratio,valign=c]{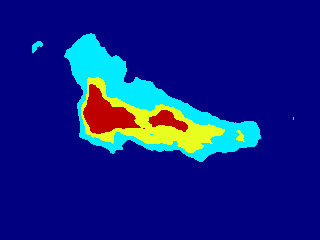} &
\includegraphics[width=0.3\textwidth, height=24mm, keepaspectratio,valign=c]{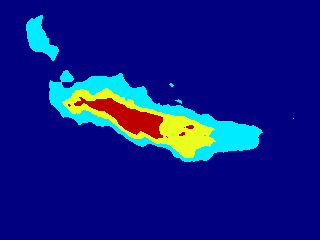} &
\includegraphics[width=0.3\textwidth, height=24mm, keepaspectratio,valign=c]{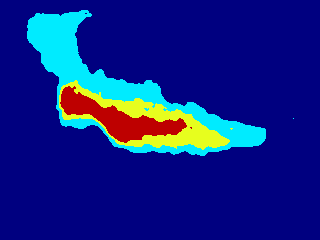} &
\includegraphics[width=0.3\textwidth, height=24mm, keepaspectratio,valign=c]{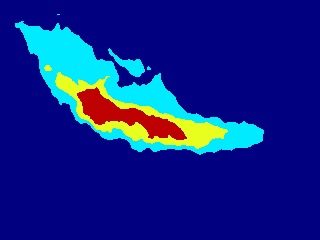} \\
\hline

Pruned UNet (Jaccard) &
\includegraphics[width=0.3\textwidth, height=24mm, keepaspectratio,valign=c]{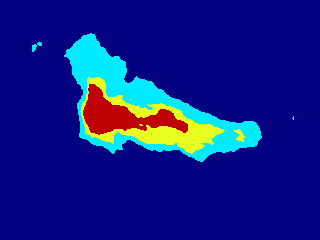} &
\includegraphics[width=0.3\textwidth, height=24mm, keepaspectratio,valign=c]{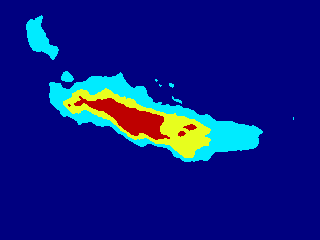} &
\includegraphics[width=0.3\textwidth, height=24mm, keepaspectratio,valign=c]{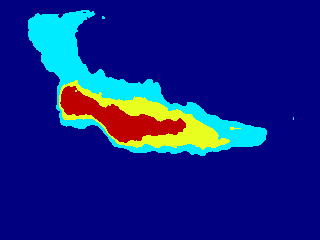} &
\includegraphics[width=0.3\textwidth, height=24mm, keepaspectratio,valign=c]{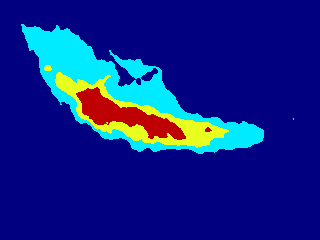} \\
\hline

UNet Batch Normalization (Cross Entropy)  &
\includegraphics[width=0.3\textwidth, height=24mm, keepaspectratio,valign=c]{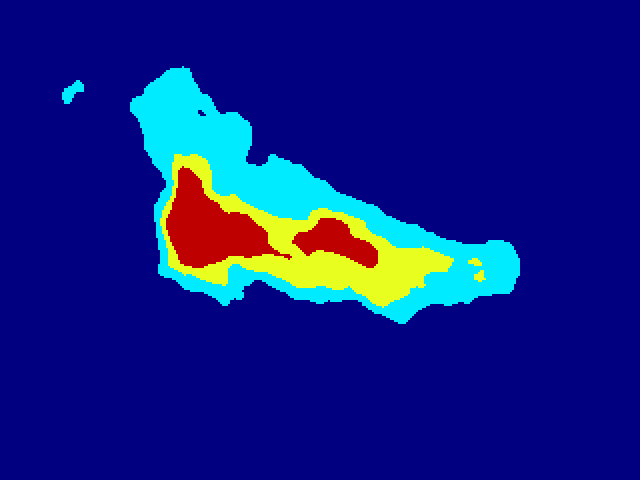} &
\includegraphics[width=0.3\textwidth, height=24mm, keepaspectratio,valign=c]{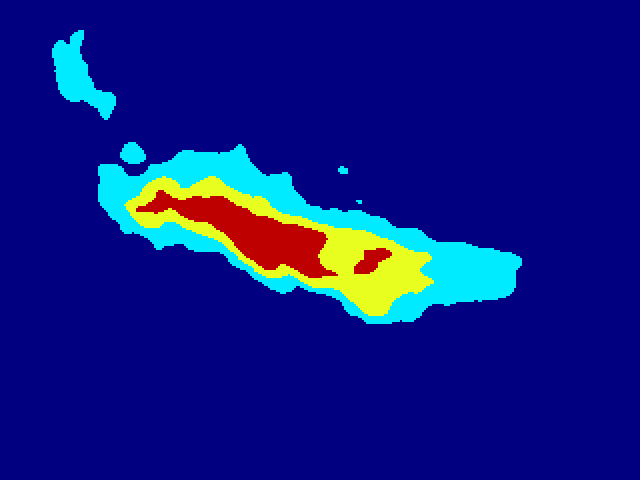} &
\includegraphics[width=0.3\textwidth, height=24mm, keepaspectratio,valign=c]{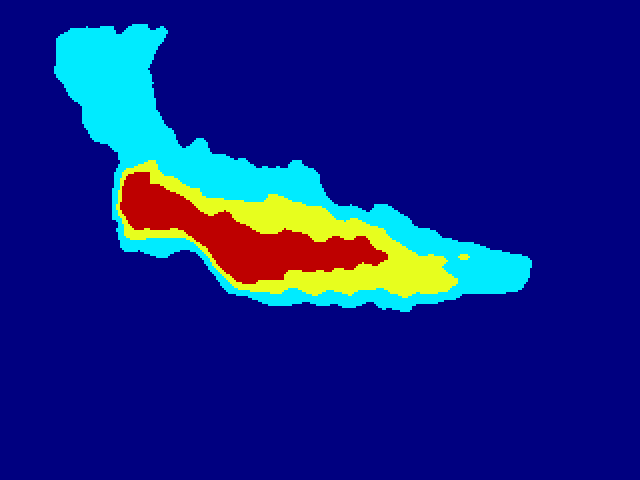} &
\includegraphics[width=0.3\textwidth, height=24mm, keepaspectratio,valign=c]{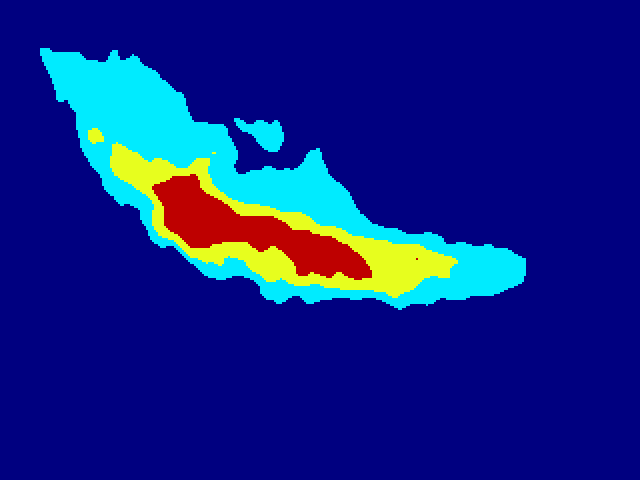} \\
\hline

Pruned UNet Batch Normalization (Cross Entropy) &
\includegraphics[width=0.3\textwidth, height=24mm, keepaspectratio,valign=c]{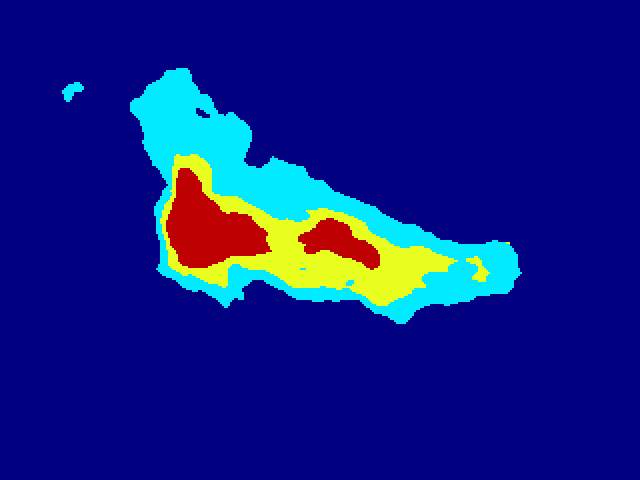} &
\includegraphics[width=0.3\textwidth, height=24mm, keepaspectratio,valign=c]{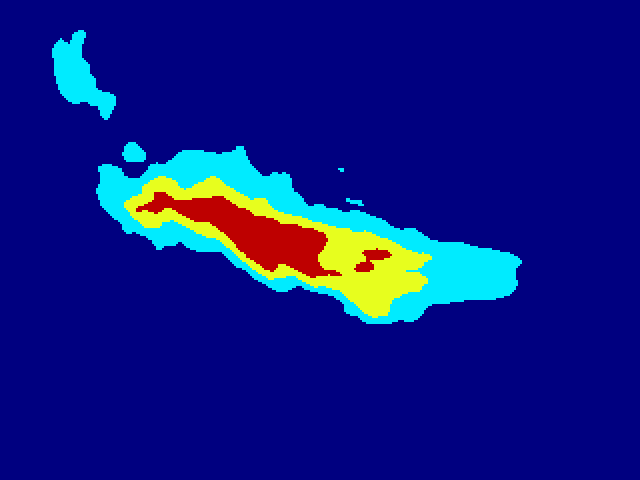} &
\includegraphics[width=0.3\textwidth, height=24mm, keepaspectratio,valign=c]{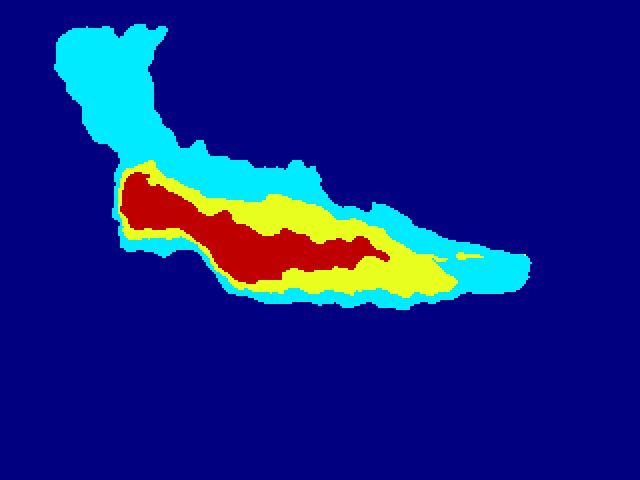} &
\includegraphics[width=0.3\textwidth, height=24mm, keepaspectratio,valign=c]{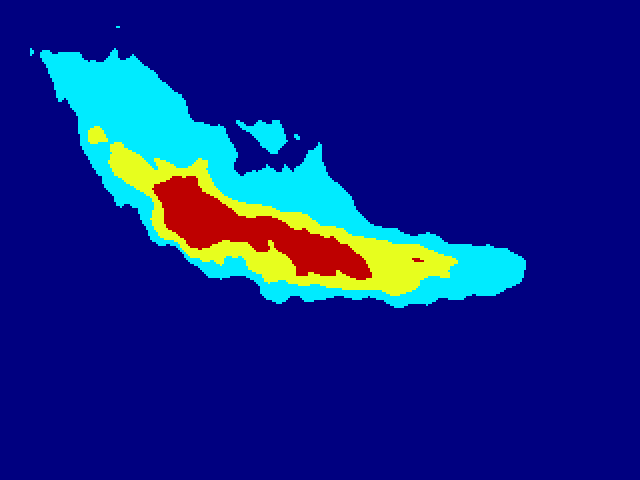} \\
\hline

\multicolumn{5}{p{440pt}}{}\\
\end{tabular}

\end{table*}









\subsection{Computational performance}
In this section, we discuss some important computational performance aspects (parameters count, number of operations, inference time) related to our model implementation for jet fire segmentation. In the previous section we showed that with the optimized model a very similar performance to the full-precision UNet implementation was obtained. 
We discuss now the computational performance improvements obtained by our FPGA-based implementation. To this end, we discuss the computational complexity of each of the blocks encompassing a UNet model using \gls{MACs} and \gls{FLOPs} values.

\begin{table*}
\caption{Computational complexity analysis. M refers to millions.}
\label{tab: complexity_analysis}
\setlength{\tabcolsep}{3pt}
\begin{tabular}{|p{205pt}|p{60pt}|p{60pt}|}
\hline
\textbf{Model (Metric)} & 
\textbf{MACs} &
\textbf{FLOPs} \\
\hline
UNet (Dice) & 48,587 M & 97,262 M\\
Pruned UNet (Dice) & 7,853 M & 15,752 M   \\
UNet (Jaccard) & 48,587 M & 97,262 M\\
Pruned UNet (Jaccard) & 8,022M & 16,091 M  \\
UNet Batch Normalization (Cross Entropy)  & 41,394 M & 82,871 M \\
Pruned UNet Batch Normalization (Cross Entropy) & \textbf{401 M} & \textbf{811 M}\\
\hline
\multicolumn{3}{p{200pt}}{}\\
\end{tabular}
\end{table*}

The results of applying these analyses to the models implemented in this paper are summarized in Table \ref{tab: complexity_analysis}. The number of required \gls{MACs} is reduced to about 17\% for the optimized implementation, and the number of \gls{FLOPs} is reduced to about the same amount. This reduction is reflected in the improved \gls{DPU} latency, which is reduced from 0.168 s to 0.059 s for the UNet (Dice) models, and from 0.164 s to 0.065 s for the UNet (Jaccard) models, as shown in the first column of Table \ref{tab: results_frame_rate}. This represents a  time reduction of about 65\% in both cases. This improvement alone is reflected on a 2.2x improvement in terms of the attainable frame rate of the evaluated models on the single-thread, and of a 2.5x improvement on the multi-thread setting (see Table \ref{tab: results_frame_rate}). On the other hand, the Pruned UNet Batch Normalization model achieved the highest speed (29.94 \gls{FPS}). This improvement is directly related to the reduction in the number of parameters, \gls{MACs}, and \gls{FLOPs}. With fewer operations required, the model exhibits lower \gls{DPU} latency, which remains the primary bottleneck in image processing tasks.

\begin{table*}[htpb!]
\caption{\gls{DPU} latency and experimental frame rate measured for every model using single- and multi-threading approaches.}
\label{tab: results_frame_rate}
\setlength{\tabcolsep}{3pt}
\begin{tabular}{|p{200pt}|p{60pt}|p{80pt}|p{60pt}|}
\hline
\textbf{Model (Metric)} & 
\textbf{DPU latency} &
\textbf{Single-thread frame rate} &
\textbf{Multi-thread frame rate}\\
\hline
UNet (Dice) & 0.168 s & 5.183 fps & 5.83 fps \\
Pruned UNet (Dice) & 0.059 s & 11.44 fps & 14.98 fps\\
UNet (Jaccard) & 0.164 s & 5.20 fps& 5.91 fps\\
Pruned UNet (Jaccard) & 0.065 s & 10.98 fps& 14.01 fps\\
UNet Batch Normalization (Cross Entropy)  & 0.185 s & 4.51 fps & 1.45 fps\\
Pruned UNet Batch Normalization (Cross Entropy)  & \textbf{0.023 s} & \textbf{14.39 fps} & \textbf{29.94 fps}\\
\hline
\multicolumn{4}{p{240pt}}{}\\
\end{tabular}
\end{table*}

\begin{table}[h]
\centering
\caption{MAPE and RMSPE errors for 3 full precision segmentation models (UNet, Attention UNet, UNet++) and the best performing optimized model (UNet with Batch Normalization (BN)). Results are reported for a nozzle of 0.02 m in diameter.}
\label{tab: results_nozzle}
\begin{tabular}{|l|l|c|c|c|}
\hline
\textbf{Model} & \textbf{Metric} & \textbf{Height} & \textbf{Area} & \textbf{Lift-Off} \\ \hline
\multirow{2}{*}{UNet} & MAPE & 2.9\% & 7.9\% & 5.4\% \\ 
 & RMSPE & 3.7\% & 9.2\% & 6.6\% \\ \hline
\multirow{2}{*}{Attention UNet} & MAPE & 2.9\% & 6.5\% & 5.2\% \\ 
 & RMSPE & 3.4\% & 8.2\% & 7.2\% \\ \hline
\multirow{2}{*}{UNet++} & MAPE & 2.9\% & 7.1\% & 5.8\% \\  
 & RMSPE & 3.6\% & 8.6\% & 7.7\% \\ \hline
\multirow{2}{*}{UNet with BN} & MAPE & 1.2\% & 5.8\% & 2.4\% \\ 
 & RMSPE & 2.8\% & 6.0\% & 2.9\% \\ \hline
\end{tabular}
\end{table}

\subsection{Jet fire geometric characterization}


After segmentation, the obtained images were analyzed to obtain: (i) the distance between the base of the stable ﬂame and the tip of the ﬂame, defined as the flame height ($L$); (ii) the distance between the pipe outlet diameter and the base of the stable ﬂame, defined as the lift-off distance ($S$); and (iii) the flame area ($A$). Obtaining these geometrical features is relevant to determine: (i) thermal radiation, which can be very high at short distances, and (ii) flame impingement on a person or nearby equipment.

\begin{figure*}[t]
    \centering
    \begin{subfigure}[t]{0.3\textwidth}
        \centering
        \includegraphics[width=\linewidth]{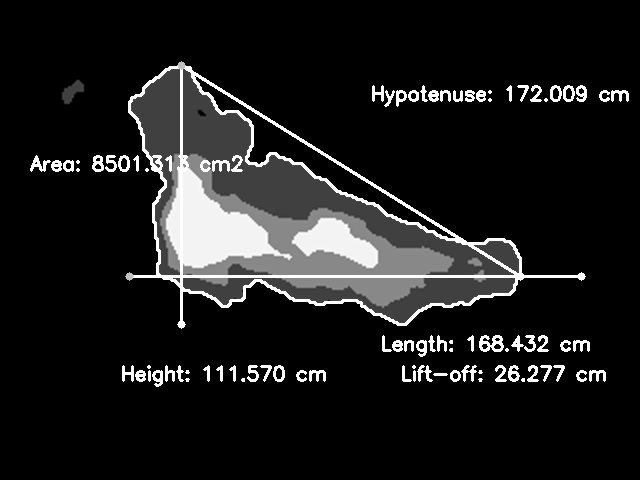}
        \caption{}
        \label{fig:sub1}
    \end{subfigure}
    \begin{subfigure}[t]{0.3\textwidth}
        \centering
        \includegraphics[width=\linewidth]{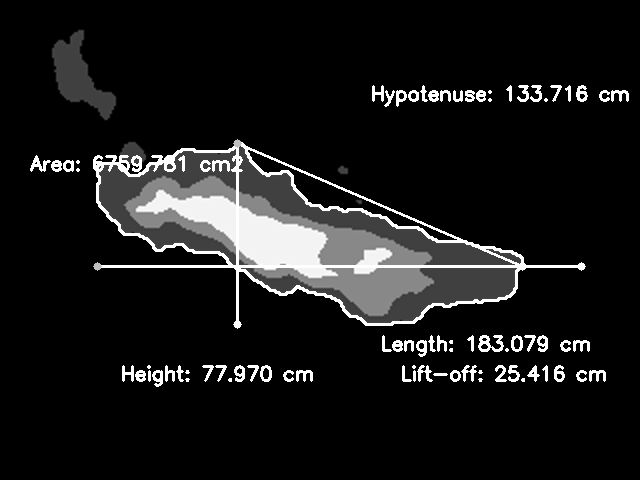}
        \caption{}
        \label{fig:sub2}
    \end{subfigure}

    \begin{subfigure}[t]{0.3\textwidth}
        \centering
        \includegraphics[width=\linewidth]{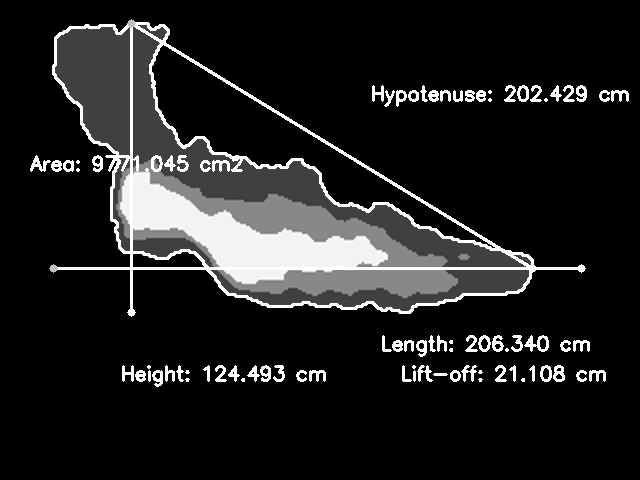}
        \caption{}
        \label{fig:sub3}
    \end{subfigure}
    \begin{subfigure}[t]{0.3\textwidth}
        \centering
        \includegraphics[width=\linewidth]{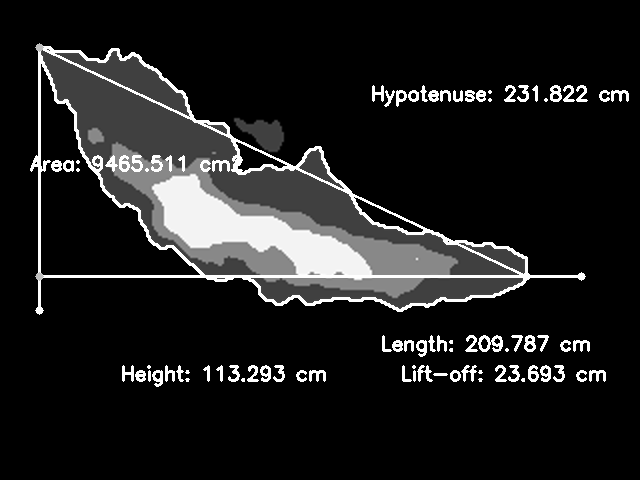}
        \caption{}
        \label{fig:sub4}
    \end{subfigure}

    \caption{Four examples of jet flame measurements using the proposed approach.} 
    \label{fig:Flame_measures}
\end{figure*}

To further analyze the results of the segmentation obtained by a selected \gls{DL} model, the jet fire's geometrical information was extracted from the generated mask and compared against the numeric experimental data. This evaluation was done using two different error measures, \gls{MAPE} and \gls{RMSPE}. As an example, in Table \ref{tab: results_nozzle} we provide results for a nozzle with a diameter of 0.02 m. This table includes the \gls{MAPE} and \gls{RMSPE} for 3 different models (full precision UNet, Attention UNet and UNet++) together with our best model (Pruned UNet Batch Normalization). We carried out these experiments with more recent versions of the UNet model, as discussed in \cite{carmina_journal} against our basic baseline, but other segmentation models could easily be implemented using our framework. From the results we can attest that the proposed model does maintain a low error but improves the overall results, as expected. In Figure \ref{fig:Flame_measures} the results obtained after running the segmentation inference and characterizing geometrically the jet flames are presented. As discussed before, this process is performed in the \gls{PS} components of the proposed platform with negligible computing overhead.

\section{Discussion}
\label{Discussion}

In this work, we extend previous research on computer-vision jet fire segmentation by proposing a novel experimental setup and system implementation for real-time fire safety applications.  The core innovation involves optimizing a UNet segmentation model to make it compatible with an \gls{SoC} \gls{FPGAs} implementation, which enables massively parallel execution of image processing tasks on device, reducing video processing overheads and overall latency.

A key achievement of this work is a significant model compression and optimization.  Using the Vitis (Xilinx) framework, we optimized the full precision model from 7.5 million parameters to 59,095 parameters (a 125x reduction), which translated into a reduction of processing latency of 2.9x. Further optimizations involving multi-threading and batch normalization led to an improvement of 7.5x in terms of latency, yielding a performance of 30 \gls{FPS} without sacrificing accuracy in terms of the evaluated metric (i.e., \gls{JI}). This is comparable to results obtained by full floating-point models running on \gls{GPU}s

 Nonetheless, several significant potential areas of opportunity for further improvement through optimization strategies are possible. The first strategy involves substituting regular convolutional layers by depth-wise separable convolution. This approach requires less computing power compared to standard convolutions.  Depth-wise separable convolution decomposes standard convolutions into depth-wise and point-wise operations, significantly reducing the number of parameters and computations required while potentially maintaining or improving model performance.

The second strategy is related to the adoption of more efficient segmentation architectures. Such efficient architectures have already been proven effective in the literature but our intent in this work was to show how any segmentation model could be integrated into an operational smart camera.  Modifying the base model would involve replacing the current UNet architecture by architectures specifically designed for computational efficiency without sacrificing segmentation quality. We plan to investigate these architectures in the future.

\section{Conclusion}
\label{Conclusion}

In the present work, we have implemented and analyzed the performance of a smart camera system based on an \gls{FPGAs} accelerator. To do so, we have used an optimized version of a UNet architecture model. We also observed that applying techniques such as quantization and pruning to an \gls{NN} model effectively reduced inference time while producing good-quality results in flame segmentation tasks of \gls{IR} images. The observed results were as good as those obtained by full floating point models running on \gls{GPU}s \citep{motiv_perez_2021}. The frame rate obtained for the segmentation task was 29.94 \gls{FPS}.

By implementing flame characterization through zones segmentation, and geometric features estimation directly on an smart camera, the system enables autonomous decision-making at the point of data capture. This implies that industrial safety systems can become more responsive and proactive, detecting and characterizing hazardous conditions instantaneously rather than relying on delayed analysis from centralized systems.

Finally, all the possibilities previously discussed and the current results make us believe that heavy computing tasks executed on the edge, such as jet fire surveillance, can benefit from accelerators implemented in configurable devices such as \gls{FPGAs}. The successful implementation on an \gls{SoC} \gls{FPGAs} (Ultra96 platform) demonstrates that industrial-grade safety systems can be deployed using configurable, cost-effective hardware. This has implications for scaling safety solutions across multiple facilities or production lines, as each deployment can operate independently without overwhelming centralized computing resources.

\printglossary[type=\acronymtype, title=Abbreviations]

\section*{Acknowledgments}
The authors wish to acknowledge the Mexican Secretaría de Ciencia, Humanidades, Tecnología e Innovación (SECIHTI) and Tecnológico de Monterrey for their support in terms of postgraduate scholarships in this project, and the Data Science Hub at Tecnológico de Monterrey for their support in this project.
This work has been supported by Azure Sponsorship credits granted by Microsoft's AI for Good Research Lab through the AI for Health program.
We also gratefully acknowledge the support from the Google Explore Computer Science Research (CSR) Program for partially funding this project through the LATAM Undergraduate Research Program. A. Àgueda is a Serra Húnter Fellow.

\bibliographystyle{cas-model2-names}
\bibliography{biblio}

\end{document}